\documentclass[lettersize,journal]{IEEEtran}
\usepackage{amsmath,amsfonts}
\usepackage{algorithmic}
\usepackage{algorithm}
\usepackage{array}
\usepackage[caption=false,font=normalsize,labelfont=sf,textfont=sf]{subfig}
\usepackage{textcomp}
\usepackage{stfloats}
\usepackage{url}
\usepackage{verbatim}
\usepackage{bm}
\usepackage{multirow}
\usepackage{graphicx}
\usepackage{cite}
\usepackage{hyperref} 
\usepackage{authblk}
\usepackage[capitalize]{cleveref}
\usepackage{subfig}
\usepackage[table]{xcolor}
\usepackage{color}
\usepackage{booktabs}
\usepackage[capitalize]{cleveref}
\usepackage[table,xcdraw]{xcolor}
\usepackage[normalem]{ulem}
\useunder{\uline}{\ul}{}
\newcommand{\ie}{\emph{i.e.}{}}
\newcommand{\eg}{\emph{e.g.}{}}

\newcommand{\equalcontrib}{\textsuperscript{*}}
\newcommand{\corrauth}{\textsuperscript{\dag}}

\newcommand{\revision}{} 
\newcommand{\delete}[1]{} 

\begin{document}

\title{Token Caching for Diffusion Transformer Acceleration}



\author{
        Jinming~Lou\equalcontrib,
        Wenyang~Luo\equalcontrib,
        Yufan~Liu\equalcontrib \corrauth,
        Bing~Li,
        Xinmiao~Ding,
        Weiming~Hu,
        ~\IEEEmembership{Senior Member,~IEEE},
        Yuming~Li,
        and~Chenguang~Ma

\thanks{
This work was supported by Beijing Natural Science Foundation (JQ24022), the National Natural Science Foundation of China (No. 62372451, No. 62192785, No. 62372082), CAAI-Ant Group Research Fund (CAAI-MYJJ 2024-02), Young Elite Scientists Sponsorship Program by CAST (2024QNRC001).
}

\thanks{
Jinming Lou and Xinmiao Ding are with the State Key Laboratory of Multimodal Artificial Intelligence Systems, Institute of Automation, Chinese Academy of Sciences, Beijing 100190, China, and the School of Information and Electronic Engineering, Shandong Technology and Business University, Yantai 264005, China (e-mail: 2022420059@sdtbu.edu.cn).}

\thanks{
Wenyang Luo, Yufan Liu, and Bing Li are with the State Key Laboratory of Multimodal Artificial Intelligence Systems, Institute of Automation, Chinese Academy of Sciences, Beijing 100190, China. Yufan Liu and Wenyang Luo are also with the School of Artificial Intelligence, University of Chinese Academy of Sciences, Beijing 100190, China. Bing Li is also with PeopleAl, Inc., Beijing 100190, China (e-mails: luowenyang2020@ia.ac.cn; yufan.liu@ia.ac.cn; bli@nlpr.ia.ac.cn).}

\thanks{
Weiming Hu is with the State Key Laboratory of Multimodal Artificial Intelligence Systems, Institute of Automation, Chinese Academy of Sciences, Beijing 100190, China; the School of Artificial Intelligence, University of Chinese Academy of Sciences, Beijing 100190, China; and the School of Information Science and Technology, ShanghaiTech University, Shanghai 201210, China (e-mail: wmhu@nlpr.ia.ac.cn).}

\thanks{
Yuming Li and Chenguang Ma are with the Terminal Technology Department, Alipay, Ant Group, China.}

\thanks{$^*$These authors contributed equally to this work.}
\thanks{\corrauth Yufan Liu is the corresponding author.}}

\markboth{Journal of \LaTeX\ Class Files,~Vol.~14, No.~8, August~2021}%
{Shell \MakeLowercase{\textit{et al.}}: A Sample Article Using IEEEtran.cls for IEEE Journals}


\maketitle

\begin{abstract}
Diffusion transformers have gained substantial interest in diffusion generative modeling due to their outstanding performance. However, their computational demands, particularly the quadratic complexity of attention mechanisms and multi-step inference processes, present substantial bottlenecks that limit their practical applications. To address these challenges, we propose TokenCache, a novel acceleration method that leverages the token-based multi-block architecture of transformers to reduce redundant computations. TokenCache tackles three critical questions: (1) Which tokens should be pruned and reused by the caching mechanism to eliminate redundancy? (2) Which blocks should be targeted for efficient caching? (3) At which time steps should caching be applied to balance speed and quality? In response to these challenges, TokenCache introduces a Cache Predictor that hierarchically addresses these issues by (1) Token pruning: assigning importance scores to each token to determine which tokens to prune and reuse; (2) Block selection: allocating pruning ratio to each block to adaptively select blocks for caching; (3) Temporal Scheduling: deciding at which time steps to apply caching strategies. Experimental results across various models demonstrate that TokenCache achieves an effective trade-off between generation quality and inference speed for diffusion transformers. 
\end{abstract}


\section{Introduction}
\label{sec:intro}

\IEEEPARstart{D}{iffusion} models \cite{ho2020denoising, song2020denoising, song2020score, rombach2022high} have established new benchmarks in generative modeling by excelling in image, video, and text generation through an iterative process of noise refinement.
Beyond their success in general generation tasks, diffusion models have also demonstrated remarkable potential in various downstream applications, including image editing \cite{saharia2022photorealistic,valevski2023unitune,zhang2024mmginpainting}, where they enable fine-grained modifications and inpainting; image restoration \cite{huang2024wavedm,chung2023parallel}, where they are leveraged for tasks such as denoising, super-resolution, and deblurring; and personalized content generation \cite{wang2023stylediffusion,everaert2023diffusion,xu2024sgdm}, where they facilitate the creation of customized outputs guided by user preferences or conditional inputs.
Recently, the Diffusion Transformer (DiT) \cite{peebles2023dit} has been proposed as a transformer-based alternative to the commonly used U-Net architectures \cite{ronneberger2015u} in diffusion models. Advancements such as SD3 \cite{esser2024scaling}, PixArt-$\alpha$ \cite{chen2023pixart}, and Sora \cite{videoworldsimulators2024}, highlight DiT's ability to effectively produce high-quality data.

\begin{figure}[t]
\centering
\hspace{-1.5em}
\includegraphics[width=0.49\textwidth]{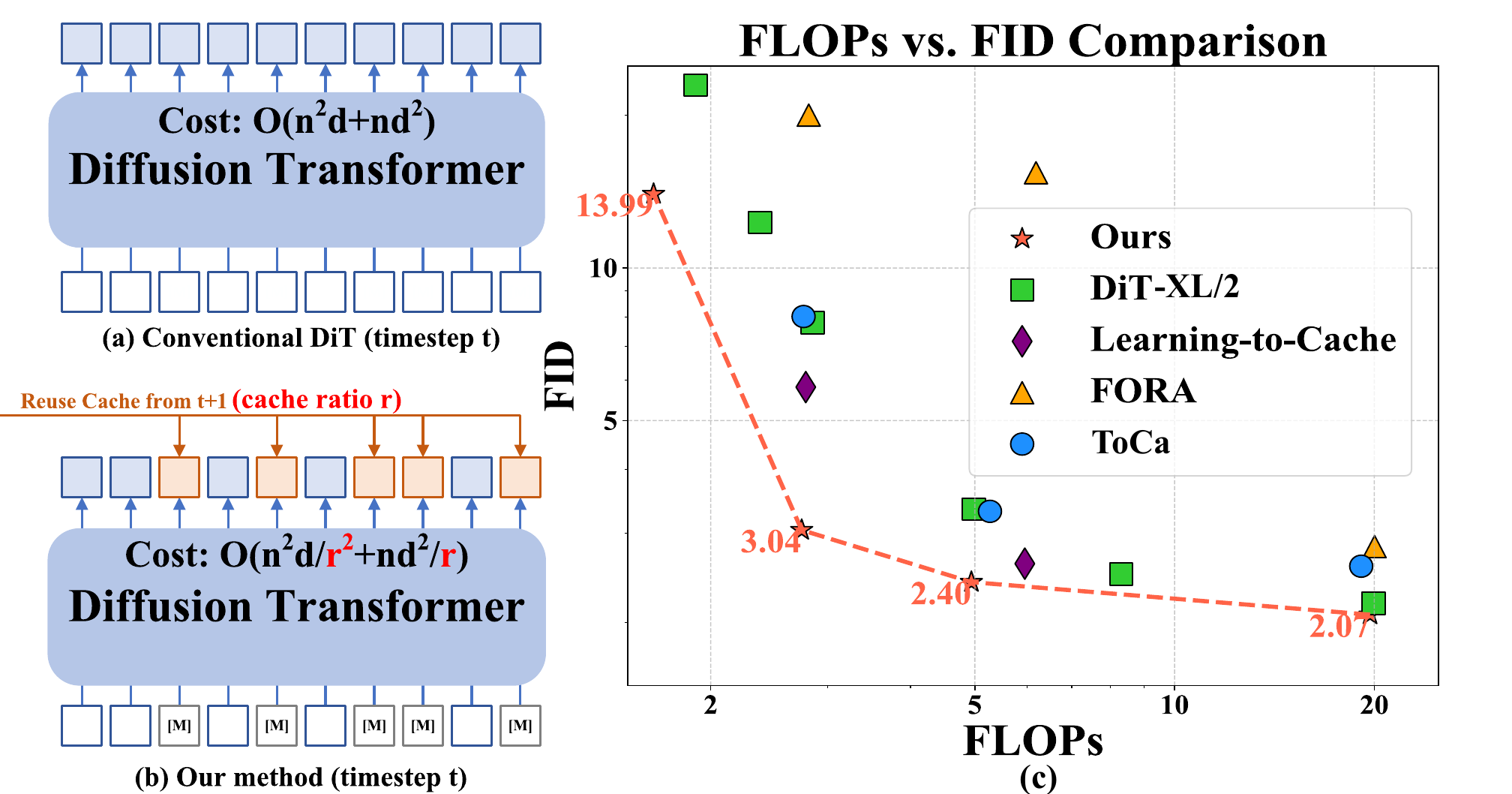}
\caption{(a) Difference between the conventional DiT model and our TokenCache method; (b) our method achieves a favorable quality-speed trade-off for accelerating DiT. Our TokenCache is superior to the compared acceleration techniques: FORA \cite{selvaraju2024fora}, Learning-to-Cache \cite{ma2024learning}, DiT-XL/2 \cite{peebles2023dit}, ToCa \cite{toca}.}
\label{fig:fig1}
\vspace{-.6em}
\end{figure}

Despite the success of diffusion models, they incur significant computational overhead that leads to slow inference speeds. Existing acceleration methods, such as improved sampling techniques \cite{song2020ddim, lu2022dpm}, consistency models \cite{luo2023latent}, quantization \cite{shang2023post, li2023q}, distillation \cite{yang2023diffusion, poole2022dreamfusion, salimans2022progressive}, and caching \cite{ma2024deepcache, li2023faster, so2023frdiff}, have primarily targeted U-Net-based diffusion models. The DiT, as a new paradigm in diffusion generative models, faces unique challenges. Beyond the inherent slowness due to multiple inference steps, it also suffers from the quadratic complexity of its attention mechanism, leading to substantial computational costs. Currently, few methods are specifically designed to exploit the unique architecture of DiT. Consequently, applying existing acceleration techniques to DiT does not yield optimal performance improvements.

Recent caching methods \cite{ma2024deepcache, li2023faster, so2023frdiff} show promise, particularly by leveraging the iterative nature of diffusion sampling processes. They store and reuse intermediate results from the U-Net blocks to reduce redundancy computations, thereby achieving a favorable trade-off between generation quality and speed. However, these methods are primarily block-level and do not consider the token-based computation characteristics of DiT. To fully tap into DiT's acceleration potential, more fine-grained acceleration strategies that align with its token-based architecture are required.

To address the above limitations, \delete{We} \revision{we} propose TokenCache, a novel acceleration method that integrates caching techniques with DiT. In developing TokenCache, we tackle three key questions: \textit{which time steps} to apply to cache, \textit{which blocks} to target, and \textit{which tokens} to cache. 

We introduce a Cache Predictor to solve the above questions hierarchically. 
To determine which tokens to cache, the Cache Predictor assigns importance scores to each token, enabling selective caching and reuse of redundant tokens without compromising model performance. \delete{The computation results of the selected tokens are replaced by their previously cached features. } \revision{We replace the computation results of the selected tokens with their previously cached features.}  
Regarding which blocks to target, the Cache Predictor adaptively allocates the cache ratio to each block.
\delete{Block importance and the corresponding allocated cache ratio are determined by aggregating the importance scores of their tokens, focusing on those blocks with the least impact on network output.} 
\revision{We determine the block importance and the corresponding cache ratio by aggregating the importance scores of their tokens, focusing on blocks that have the least impact on the network output.}
Finally, the Cache Predictor introduces temporal scheduling to determine which time steps to apply caching. Based on the importance scores, time steps are divided into independent inference steps (I-steps) and cached prediction steps (P-steps). \delete{Full computations are required in I-steps because the details and information are crucial for the generative quality. At P-steps, efficient caching mechanisms are employed to reduce redundancy and enhance speed.}
\revision{We perform full computations in I-steps because detailed information is crucial for maintaining generative quality, while at P-steps we employ efficient caching mechanisms to reduce redundancy and enhance speed.}

As illustrated in \cref{fig:fig1}, our TokenCache method significantly enhances the conventional DiT model by achieving an optimal balance between generation quality and inference speed. 
Experimental results across various models demonstrate that TokenCache achieves an effective trade-off between generation quality and inference speed for DiT.

Our contributions are summarized as follows:
\begin{itemize}
    \item We present a novel acceleration framework for DiT based on caching and reuse of intermediate tokens. To the best of our knowledge, this is one of the first attempts to accelerate DiT at the token level.
    \item We introduce a Cache Predictor that assigns importance scores to tokens, enabling selective caching of redundant tokens without compromising model performance. This reduces computational redundancy by reusing cached token results.
    \item The proposed Cache Predictor allocates cache ratios to each block, adaptively selecting blocks for caching and reuse. Besides, it also implements a temporal scheduling mechanism that divides timesteps into I-steps and P-steps, which optimizes caching intervals across the denoising process to balance acceleration with minimal quality loss.
\end{itemize}

\begin{figure*}[h]
\centering
\includegraphics[width=\textwidth]{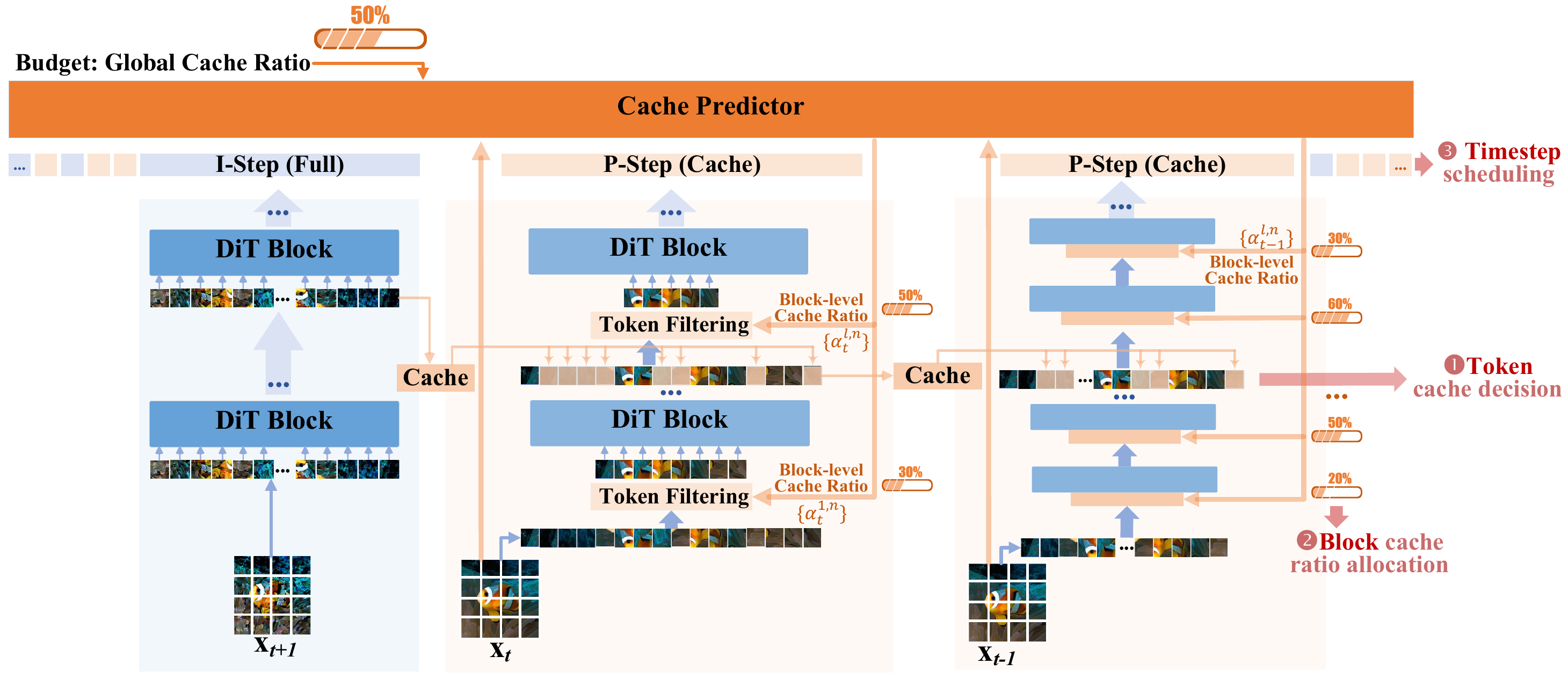} 
\caption{Framework of TokenCache. TokenCache decomposes the space of caching strategies into three dimensions: 1) Which tokens to cache: we propose the Cache Predictor to estimate the importance of tokens, enabling the pruning and reusing of cached values; 2) Which blocks to cache tokens: we allocate caching ratio to each block to select blocks for caching adaptively; 3) Which timesteps to perform caching: we introduce an adaptive timestep schedule that interleaves non-caching I-steps and caching P-steps.}
\label{fig:framework}
\vspace{-.6em}
\end{figure*}

\section{Related Work}

\subsection{Diffusion Models}

Diffusion models \cite{ho2020denoising, song2020denoising, song2020score, rombach2022high} generate data $\bm{x}_0$ by reversing a Markov chain that gradually adds Gaussian noise $\mathcal{N}$ to data. At each timestep $t=0, \dots, T$ where $T$ is the number of timesteps, the forward process is defined by
\begin{equation}
q(\bm{x}_t \mid \bm{x}_{t-1}) = \mathcal{N}(\bm{x}_t; \sqrt{\alpha_t}\bm{x}_{t-1}, (1-\alpha_t)\bm{I}),
\end{equation}
where $\alpha_t$ is the noise level.
The reverse process correspondingly denoises $\bm{x}_t$ by predicting the noise component $\bm{\epsilon}_{\theta}(\bm{x}_t, t)$ and estimating $\bm{x}_{t-1}$ at the previous timestep:
\begin{equation}
\bm{x}_{t-1} = \frac{1}{\sqrt{\alpha_t}} \left( \bm{x}_t - \frac{1-\alpha_t}{\sqrt{1-\bar{\alpha}_t}} \bm{\epsilon}_{\theta}(\bm{z}_x, t) \right) + \sigma_t \bm{\epsilon},
\label{eq:backward}
\end{equation}
with $\bm{\epsilon} \sim \mathcal{N}(0, \bm{I})$. Iterating \cref{eq:backward} from $t=T$ to $t=0$ generates the data $\bm{x}_0$ from Gaussian noise $\bm{x}_{T} \sim \mathcal{N}(0, \bm{I})$.

DiT \cite{peebles2023dit} introduces transformers \cite{vaswani2017attention, dosovitskiy2020image} to diffusion models and demonstrates better scaling properties than U-Nets \cite{ronneberger2015u}. Recent advances like PixArt-$\alpha$ \cite{chen2023pixart}, SD3 \cite{esser2024scaling}, and Sora \cite{videoworldsimulators2024} showcase DiT's scalable capabilities in generating high-quality data, motivating our exploration into DiT.

\subsection{Efficient and Accelerated Diffusion}

Diffusion models \cite{ho2020denoising, song2020denoising, song2020score, rombach2022high} have demonstrated impressive generation quality but often suffer from slow sampling speeds and high computational costs. Recent research on accelerating diffusion models can be broadly grouped into two main directions: reducing the number of sampling steps and compressing or restructuring the denoising networks. In the first direction, reducing Sampling Steps, various sampler-based techniques focus on improving solver efficiency or decreasing the required iterations for generating high-quality samples. For instance, DDIM \cite{song2020ddim} introduces a non-Markovian sampling strategy that reduces timesteps without sacrificing image fidelity. DPM-Solver \cite{lu2022dpm}, DPM-Solver++ \cite{lu2022dpm++}, and EDM \cite{karras2022elucidating} propose advanced numerical schemes to solve ordinary differential equations (ODEs) and stochastic differential equations (SDEs) more efficiently, enabling fewer steps while maintaining generation quality.

Another approach targets network-side acceleration, such as pruning, quantization, and knowledge distillation \cite{salimans2022progressive, luo2023latent, shang2023post, li2023q, yang2023diffusion, poole2022dreamfusion}. These techniques aim to reduce model complexity or parameter sizes. Progressive Distillation \cite{salimans2022progressive} accelerates the sampling of diffusion models by progressively distilling a teacher model into a student model, reducing the number of required sampling steps while maintaining high sample quality. LCM \cite{luo2023latent} optimizes noise estimation networks based on timestep dependencies, allowing the model to produce samples in fewer iterations. Despite these improvements, they often involve complex training processes or may not fully optimize the inference phase.

In addition, diffusion inference can be further accelerated by using cache technology, which will be discussed in detail in the next section. This approach leverages previously computed results to skip redundant calculations, offering an efficient way to speed up the generation process. However, implementing effective caching strategies requires careful consideration of when and what to cache, as improper use can lead to negligible benefits or even performance degradation.

\subsection{Caching for Diffusion Acceleration}

The multi-step iterative generation process (\cref{eq:backward}) of diffusion models leads to high inference costs. However, since the intermediate $\bm{x}_t$ across consecutive timesteps are often similar, caching previously computed features and skipping redundant computations can accelerate inference.

Most existing caching techniques focus on the U-Net architectures.
DeepCache \cite{ma2024deepcache} caches features from deep layers because deep layers usually encode high-level features that change slowly.
Faster Diffusion \cite{li2023faster} caches features from the U-Net encoder for later reuse in the decoder. 
FRDiff \cite{so2023frdiff} reuses the residual updates to features.
T-Gate \cite{tgate} increases efficiency by caching and reusing converged cross-attention outcomes, thereby eliminating redundant cross-attention computations in the fidelity-improving stage.

DiT is a homogeneous token-based architecture distinct from the encoder-decoder structure of U-Nets, necessitating new techniques for acceleration. 
Recently, some work has explored handcrafted block-level caching strategies for DiT \cite{selvaraju2024fora, chen2024delta, ma2024learning}. For example, FORA \cite{selvaraju2024fora} statically caches all layer features across multiple timesteps, which achieves higher speedups but at the cost of significant performance loss. $\Delta$-DiT \cite{chen2024delta} caches deviations between feature maps rather than traditional feature maps.
Learning-to-Cache \cite{ma2024learning} learns static caching routers to control which layers are cached for reuse. It requires retraining the routers when the number of inference steps changes. The learned router is fixed and cannot adapt to varying input data.
These methods focus on caching layer features and lack flexibility at a more fine-grained token level. ToCa \cite{toca} further considers the layers and token scores to select suitable tokens for caching. However, it requires manual adjustment of dozens of hyperparameters, reducing its generalizability. 
In contrast, our method introduces a Cache Predictor that dynamically schedules token-level caching based on current inputs and timesteps. This approach allows for more flexible and adaptive caching decisions, optimizing the balance between generation quality and speed. Besides, it also eliminates the need for manual hyperparameter tuning and adapts effectively to different inputs, enhancing both practicality and performance.

\section{Method}
\label{sec:method}

We present TokenCache, a novel method designed to accelerate generation by caching and reusing intermediate tokens from DiT \cite{peebles2023dit}. In this section, the \textbf{framework}
of the proposed method is introduced in \cref{sec:framework}. Then, we describe the redundancy of updates to the token embeddings across inference timesteps in \cref{sec:token_caching}, which \textbf{motivates TokenCache}. \cref{sec:cache_prediction} details the design and training of the \delete{ptoposed} \revision{proposed} \textbf{Cache Predictor}. \cref{sec:inference} details the \textbf{inference} of TokenCache.


\subsection{Framework}
\label{sec:framework}
\cref{fig:framework} summarizes the framework of TokenCache. TokenCache operates at three levels, \ie, timestep, block, and token, to adaptively cache tokens for balanced performance and acceleration. At the timestep level, the Cache Predictor decides whether to perform full computations (I-step) or use cached tokens (P-step) at each timestep. At the block level, for timesteps identified as P-steps, the Cache Predictor allocates different cache ratios to each block based on a total cache ratio budget. This allocation ensures that blocks deemed more critical receive higher computation, while less important blocks benefit from caching strategies. \delete{At the token level, tokens are evaluated for their importance (cache decision variables) using the Cache Predictor, within each block.} \revision{At the token level, we evaluate each token’s importance (cache decision variable) using the Cache Predictor within each block.} Tokens are ranked according to these scores, and low-ranked tokens are filtered out. Instead of recomputing representations for these tokens, their values from the previous timestep are reused.

Only the high-ranked tokens undergo DiT Block computations, significantly reducing the overall computational load. This selective update mechanism allows for efficient resource utilization by focusing on the most impactful tokens. 
\delete{The entire adaptive caching strategy is managed by ONE Cache Predictor, ensuring cache decision consistency and efficient computation across all levels.}
\revision{We manage the entire adaptive caching strategy with a single Cache Predictor, ensuring consistent cache decisions and efficient computation across all levels.}
 

\begin{figure*}[t]
\centering
\includegraphics[width=\textwidth]{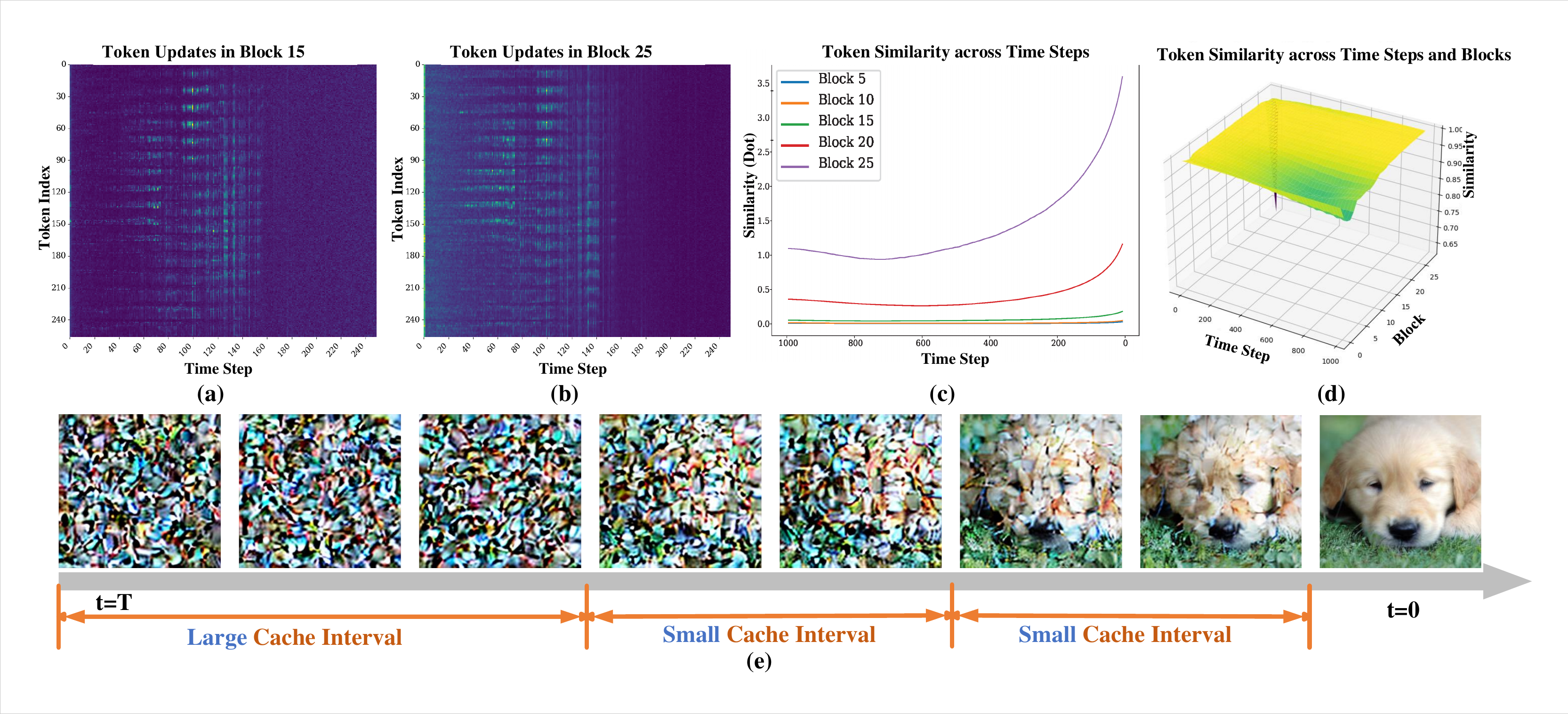} 
\caption{Demonstration of token redundancy in DiT. (a) and (b) show the heatmaps of token updates in some network blocks. (c) and (d) plot the similarity between token updates at consecutive timesteps for different blocks. Each curve in (c) represents a different block over timesteps. (e) visualizes a multi-phase timestep schedule based on the above findings.}
\label{fig:analysis}
\end{figure*}

\subsection{Token Caching for DiT}
\label{sec:token_caching}

DiT treats the input (latent \cite{kingma2013auto}) image $\bm{x}_t$ at timestep $t$ as a grid of non-overlapping patches \cite{dosovitskiy2020image} $\bm{x}_t = [\bm{x}_t^{0,1}, \bm{x}_t^{0,2}, \dots, \bm{x}_t^{0,N}]$, where patch $\bm{x}_{t}^{0,n} (1\le n \le N)$ is called an image token ($N$ is the number of tokens, $0$ indicates the index of the input layer). A stack of $L$ homogeneous transformer blocks \cite{vaswani2017attention} $\{\bm{f}^l\}_{l=1}^L$ processes the image tokens after they are embedded as high-dimensional vectors, each block consists of self-attention and Multilayer Perceptron (MLP) layers, and $\bm{f}_t^{l,n}(\bm{x}_t^{l-1})$ is the \textbf{token update} to the image token at timestep $t$ and block $l-1$:
\begin{align}
    \bm{x}_t^{l} &= [\bm{x}_t^{l,1}, \bm{x}_t^{l,2}, \dots, \bm{x}_t^{l,N}], \quad 0 \le l \le L, \label{eq:dit_1} \\
    \bm{f}_t^l &= [\bm{f}_t^{l,1}, \bm{f}_t^{l,2}, \dots, \bm{f}_t^{l,n}], \quad 0 \le l \le L, \label{eq:dit_2} \\
    \bm{x}_t^{l,n} &= \bm{x}_t^{l-1,n} + \underbrace{\bm{f}_t^{l,n}(\bm{x}_t^{l-1})}_\text{token update}, \, 1 \le l \le L, \forall n.
    \label{eq:dit_3}
\end{align}
In diffusion-based generative modeling, DiT is repeatedly applied to denoise the $\bm{x}_t$ by \cref{eq:dit_1,eq:dit_2,eq:dit_3}, starting from Gaussian noise at $t=T$.

We observe several intriguing phenomena in this process as shown in \cref{fig:analysis}:
\begin{enumerate}
    \item \textbf{Similar token update patterns across neighboring timesteps}: Token updates have similar patterns across neighboring timesteps, as seen from the horizontal strips in \cref{fig:analysis}(a) and (b). This is also evidenced by the high correlation of token updates among timesteps in various DiT blocks in \cref{fig:analysis}(c) and (d).
    \item \textbf{Different update patterns for different tokens}: Different tokens have distinct update patterns, as indicated by the different heatmap patterns and similarity distribution of updates in \cref{fig:analysis}.
    \item \textbf{Non-uniform token updates in long term}: \cref{fig:analysis}(a) and \cref{fig:analysis}(b) illustrate that token updates are not uniformly distributed in the long term. This observation necessitates designing specialized timestep scheduling strategies, such as the one demonstrated in \cref{fig:analysis}(e).
\end{enumerate}
These phenomena suggest that \textit{token updates have redundancy, but their patterns are token-specific}. This motivates us to accelerate DiT inference by skipping the updates of some tokens and reusing the cached updates of the skipped tokens from previous timesteps. 

Specifically, let $\alpha_t^{l,n} \in \{0, 1\}$ be the \textbf{cache decision variable} indicating whether the $n$-th token at timestep $t$, layer $l$ should be updated by $\bm{f}^l$ (when $\alpha_t^{l,n}=1$) or directly reused from its previously computed value at timestep $t+1$ (when $\alpha_t^{l,n}=0$). The DiT with token caching can be written as (denote $\hat{\bm{x}}_t^0 = \bm{x}_t^0$):
\begin{equation}
\begin{split}
    \hat{\bm{x}}_t^{l,n} &= \hat{\bm{x}}_t^{l-1,n} + \alpha_t^{l,n} \cdot \bm{f}_t^{l,n}(\hat{\bm{x}}_t^{l-1}) \\
    &+ (1 - \alpha_t^{l,n}) \cdot \underbrace{\bm{f}_{t-1}^{l,n}(\hat{\bm{x}}_{t-1}^{l-1})}_\text{cached update from $t-1$}, \quad 1 \le l \le L.
    \label{eq:dit_cache}
\end{split}
\end{equation}
The total computation cost of token-caching DiT (\cref{eq:dit_cache}) can be roughly estimated as
\begin{equation}
    C_\text{total} = \sum_{l,n} \alpha_t^{l,n} \times C_\text{token},
    \label{eq:cost}
\end{equation}
where $C_\text{token}$ is the average computation cost of a token through a block. 

\subsection{Cache Prediction}
\label{sec:cache_prediction}

\begin{figure*}[ht]
\centering
\includegraphics[width=.8\textwidth]{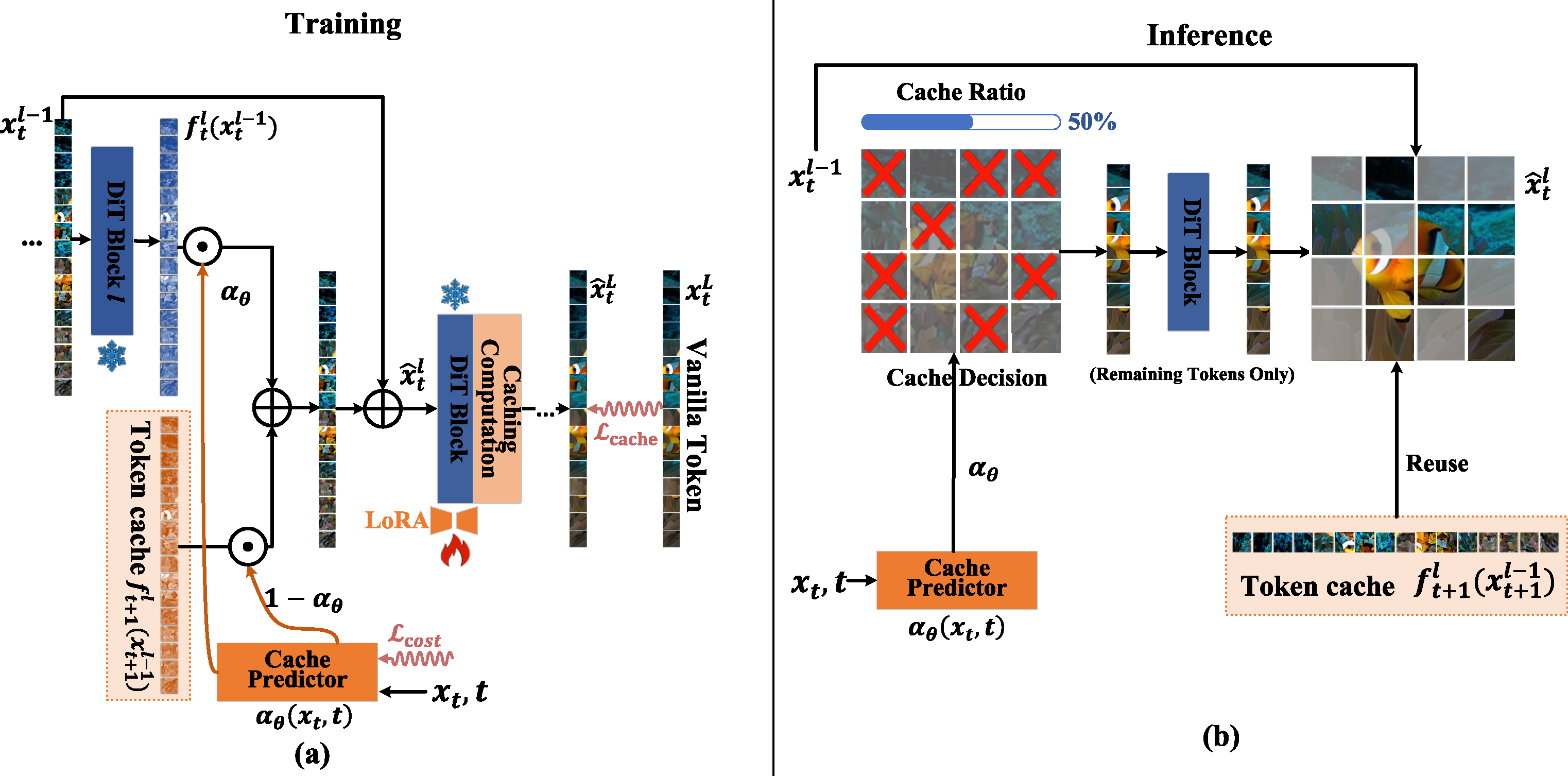}
\vspace{-.6em}
\caption{Illustration of our token caching strategy via Cache Predictor. (a) Training process, where the non-caching and caching states of the tokens are interpolated by $\bm{\alpha}_\theta$ and $1-\bm{\alpha}_\theta$. (b) Inference process. }
\label{fig:CachePredictor}
\end{figure*}

Given the global computing budget $C_\text{total}$, TokenCache aims to find the optimal combination of cache decision variables $\{\alpha_t^{l, n}\}_{l,n}$ (\ie, cache strategy) that achieves the minimal generation error.

\vspace{0.5em}
\noindent\textbf{The Design of the Cache Predictor.}
The token-level redundancy discussed in \cref{sec:token_caching} depends on both the tokens and the current timestep, as indicated by the token update heatmaps in \cref{fig:analysis}. Thus, it is sub-optimal to fix the decision variables $\alpha_t^{l, n}$ for given $t,l,n$. Instead, we choose to estimate $\{\alpha_t^{l, n}\}_{l,n}$ with a \textit{shared} time-conditioned light-weight Cache Predictor:
\begin{equation}
    \{\alpha_t^{l, n}\}_{l,n} = \bm{\alpha}_\theta(\bm{x}_t, t).
    \label{eq:cache_predictor}
\end{equation}
Given the input image $\bm{x}_t$ at timestep $t$, the Cache Predictor $\bm{\alpha}_\theta$ predicts \textit{all} decision variables in this timestep \textit{at once} as shown in \cref{fig:framework}. This allows adaptive caching based on time and image inputs with minimal computational overhead.

Specifically, the Cache Predictor $\bm{\alpha}_\theta$ is implemented as a \textit{shared} DiT block initialized with the weights of the first block. It takes $\bm{x}_t$ and $t$ as inputs, while other conditions like class labels or text embeddings remain unchanged. \delete{The output linear projection layer of the Cache Predictor is modified to produce one scalar per block and token, resulting in a total of $LN$ values.} \revision{We modify the output linear projection layer of the Cache Predictor to produce one scalar per block and token, resulting in a total of $LN$ values.} A sigmoid activation is then applied to these values to constrain them between 0 and 1. In total, the Cache Predictor adds $\sim3.57\%$ to the model's parameter count and FLOPs, ensuring efficient computation without significant additional cost.

\begin{algorithm}[h]
  \caption{\revision{Training Cache Predictor}}
  \label{alg:cachepredictor_train}
  \small
  \revision{
  \begin{algorithmic}[1]
    \STATE \textbf{Inputs:} data $p(\mathbf{x}_0)$, diffusion model $\epsilon_\theta$ (frozen), Cache Predictor $\alpha_\theta$, solver $\Psi(\cdot)$, total steps $T$
    \REPEAT
      \STATE Sample $\mathbf{x}_0 \sim p(\mathbf{x}_0)$
      \STATE Sample $t \sim \mathcal{U}\{0,\dots,T-1\}$ \textcolor{gray}{\em // train on pair $(t{+}1 \!\rightarrow\! t)$}
      \STATE Draw forward sample $\mathbf{x}_{t+1} \sim q(\mathbf{x}_{t+1}\mid \mathbf{x}_0)$
      \STATE \textbf{Teacher at $(t{+}1)$:} run $\epsilon_\theta(\mathbf{x}_{t+1}, t{+}1)$ to obtain posterior $(\mu_{t+1}, \sigma_{t+1})$
      \STATE Construct input at $t$: $\mathbf{x}_t \gets \mu_{t+1} + \sigma_{t+1}\,\epsilon$, \ $\epsilon \sim \mathcal{N}(0,I)$
      \STATE \textbf{Full path (no cache):} propagate $\mathbf{x}_t$ through $L$ blocks to get $\mathbf{x}^L_t$
      \STATE \hspace{0.9em}\textcolor{gray}{\em // per layer $\ell$: $\mathbf{x}^\ell_t=\mathbf{x}^{\ell-1}_t+f_t^\ell(\mathbf{x}^{\ell-1}_t)$}
      \STATE \textbf{Cache decisions at $t$:} $\{\alpha^{\ell,n}_{t}\} \gets \alpha_\theta(\mathbf{x}_{t}, t)$
      \STATE \textbf{Cached path:} propagate with relaxed interpolation using cached $(t{+}1)$ states
      \STATE \hspace{0.9em}\textcolor{gray}{\em // for each layer $\ell=1..L$, token $n$:}
      \STATE \hspace{0.9em}$\hat{\mathbf{x}}^{\ell,n}_t \gets \mathbf{x}^{\ell-1,n}_t
        + \alpha^{\ell,n}_{t}\, f^{\ell,n}_t(\mathbf{x}^{\ell-1}_t)
        + \big(1-\alpha^{\ell,n}_{t}\big)\, f^{\ell,n}_{t+1}(\mathbf{x}^{\ell-1}_{t+1})$
      \STATE Obtain $\hat{\mathbf{x}}^L_t$ after the last layer
      \STATE \textbf{Loss:}
      \STATE \hspace{0.9em}$\mathcal{L}_{\text{cache}} \gets \|\hat{\mathbf{x}}^L_t - \mathbf{x}^L_t\|_2^2$ \textcolor{gray}{\em // cache loss}
      \STATE \hspace{0.9em}$\mathcal{L}_{\text{cost}} \gets \|\alpha_\theta(\mathbf{x}_{t}, t)\|_1$ \textcolor{gray}{\em // cost loss on current-step gating}
      \STATE \hspace{0.9em}$\mathcal{L} \gets \mathcal{L}_{\text{cache}} + \lambda\,\mathcal{L}_{\text{cost}}$ 
      \STATE Update parameters of $\alpha_\theta$; keep $\epsilon_\theta$ frozen
    \UNTIL{converged}
  \end{algorithmic}
  }
\end{algorithm}

\vspace{0.5em}
\noindent\textbf{Optimization of the Cache Predictor.}
\revision{To make the training procedure more explicit, we provide Alg.~\ref{alg:cachepredictor_train}, 
which summarizes the Cache Predictor optimization pipeline, including the inputs, the supervision from the frozen diffusion model, and the computation of cache and cost losses.}
The goal of the Cache Predictor is to minimize the global error caused by token caching. This error is the difference between $\hat{\bm{x}}_0$ using cache strategy $\{\alpha_t^{l, n}\}_{l,n}$ and the ``ground-truth'' $\bm{x}_0$ without caching. However, estimating global errors is both costly and unstable as it requires integrating the backward diffusion process. Therefore, we focus on minimizing \textit{local errors} instead. 
Local errors are defined as the discrepancy between cached predictions $\hat{\bm{x}}_t^L$ (using \cref{eq:dit_cache,eq:cache_predictor}) and non-cached predictions $\bm{x}_t^L$ (using \cref{eq:dit_1,eq:dit_2,eq:dit_3}), starting from a ground-truth $\bm{x}_{t+1}$. The loss function is written as:
\begin{equation}
    \mathcal{L} = \mathbb{E}_{t, \bm{x}_{t+1}} [ \underbrace{\| \hat{\bm{x}}_t^L - \bm{x}_t^L \|_2^2}_\text{cache loss $\mathcal{L}_\text{cache}$}  + \lambda \underbrace{\| \bm{\alpha}_\theta(\bm{x}_{t}, t) \|_1}_\text{cost loss $\mathcal{L}_\text{cost}$} ],
    \label{eq:loss}
\end{equation}
where $\|\cdot\|_1$ is $\ell^1$ norm. The cache used at $t$ is filled by an additional model forward at $t+1$. Here, the cache loss $\mathcal{L}_\text{cache}$ minimizes the local errors caused by caching, and also limits the upper bound of global errors under mild continuous and consistency conditions \cite{iserles1993global}; the cost loss $\mathcal{L}_\text{cost}$ minimizes the computing cost $C_\text{total}$ (\cref{eq:cost}), assuming $C_\text{token}$ is constant. Losses are balanced by the parameter $\lambda > 0$.

\delete{To optimize $\bm{\alpha}_{\theta}$ via gradient descent, $\bm{\alpha}_\theta$ is relaxed to real values in $[0, 1]$.} \revision{We optimize $\bm{\alpha}_{\theta}$ via gradient descent by relaxing it to real values in $[0, 1]$.} Then, the layer output $\hat{\bm{x}}_t^{l}$ is interpolated between the current and previous timestep outputs using $\bm{\alpha}_\theta$, 
as shown in \cref{fig:CachePredictor}(a):
\begin{equation}
    \begin{split}
    \hat{\bm{x}}_t^{l,n} = \bm{x}_t^{l-1,n} &+ \bm{\alpha}_\theta^{l, n}(\bm{x}_t, t) \cdot \bm{f}_t^{l,n}(\bm{x}_t^{l-1}) \\
    &+ (1 - \bm{\alpha}_\theta^{l, n}(\bm{x}_t, t)) \cdot \bm{f}_{t+1}^{l,n}(\bm{x}_{t+1}^{l-1}).
\end{split}
\end{equation}
In addition, we fix the last $K$ layers to non-caching mode and adapt their weights with LoRA \cite{hu2021lora}, when optimizing \cref{eq:loss}. These layers act as ``refiners'' of the cached outputs to enhance consistency and overall performance.

\vspace{0.5em}
\noindent\textbf{Understanding $\bm{\alpha}$.}
The optimal solution to \cref{eq:loss}, given $\bm{x}$, is approximately as:
\begin{equation}
    \alpha_t^{l, n} \approx 1 - \lambda \left(\frac{\partial \bm{x}_t}{\partial \bm{x}_t^{l,n}} \cdot d^2(\bm{f}_{t+1}^{l,n}(\bm{x}_{t+1}^{l-1,n}), \bm{f}_t^{l,n}(\bm{x}_{t}^{l-1,n})) \right)^{-1},
    \label{eq:optimal_alpha}
\end{equation}
where $d$ is Euclidean distance.
According to \cref{eq:optimal_alpha}, the optimized $\bm{\alpha}$ reflects the similarity between token updates across timesteps, weighted by the sensitivity of local errors upon the tokens. Specifically, tokens are reused (\ie, have a small $\alpha$) when they have a low influence on errors or exhibit similar updates across time. \delete{In this context, $\bm{\alpha}$ can be regarded as an ``importance score'' of the tokens that guides our caching strategy. } \revision{In this context, we interpret $\bm{\alpha}$ as an “importance score” of the tokens that guides our caching strategy.}

\subsection{Inference with Token Caching}
\label{sec:inference}

\noindent\textbf{Token-level cache scheduling.}
During inference, TokenCache dynamically decides which tokens to cache at any timestep $t$ for a given input $\bm{x}_t$, based on the predicted cache decision variables of $\{\alpha_t^{l, n}\}_{t,l,n}$. Specifically, with a global cache ratio $r \in [0, 1]$, $\{\alpha_t^{l, n}\}_{l,n}$ is computed \textit{once} per timestep $t$ via the Cache Predictor $\bm{\alpha}_\theta$ (see Eq. \ref{eq:cache_predictor}). \delete{These cache decision variables are then quantized by setting the top $(1-r)LN$ values to $1$ (indicating these tokens should be updated) while the other values are set to $0$ (indicating reuse from the cache).}
\revision{We then quantize the cache decision variables by setting the top $(1-r)LN$ values to $1$ (indicating that these tokens should be updated) and the remaining ones to $0$ (indicating reuse from the cache).}
Tokens are subsequently updated according to \cref{eq:dit_1,eq:dit_2,eq:dit_3}.
The first timestep $t=T$ always performs full updates to initialize the cache.

\vspace{0.5em}
\noindent\textbf{Block-level cache scheduling.}
In batched inference, all samples within a batch must share the same cache ratio to ensure efficient parallel computation. To achieve this while accommodating varying token importance across different blocks, we introduce a block-level cache ratio allocation strategy. 
The core of our block-level token caching schedule is to dynamically allocate cache ratios based on the cache decision variable $\bm{\alpha}$ which reflects the token importance. Initially, we globally quantize these scores across all blocks and tokens to produce binary cache decision variables. These variables indicate whether each token should be updated ($\alpha_t^{l, n}=1$) or cached ($\alpha_t^{l, n}=0$), according to a predefined global cache ratio. Subsequently, for each individual block $l$, we compute the block-level cache ratio $r_l$ as the proportion of tokens marked for update within that block.

Formally, let $\bm{q}_l$ denote the quantized binary cache decision variables for tokens within block $l$. The block-level cache ratio $r_l$ is computed as:
\begin{equation}
r_l = \frac{\sum_{q \in \bm{q}_l} q}{|\bm{q}_l|},
\end{equation}
where $q \in \{0,1\}$ indicates whether a token is updated ($q=1$) or cached ($q=0$).

After obtaining the block-level cache ratios $r_l$, we refine the selection of tokens to cache within each block. Specifically, we re-quantize the original importance scores $\bm{\alpha}_l$ within block $l$ by retaining the top-ranked tokens according to the ratio $r_l$. This means each block dynamically decides its subset of tokens to update, reflecting their relative importance. A higher cache ratio $r_l$ indicates that more tokens within the block are cached suggesting lower importance, while a lower cache ratio signifies less caching, indicating higher importance.


\vspace{0.5em}
\noindent\textbf{Timestep scheduling.}
At the timestep level, token caching during inference can introduce local errors that accumulate over time, potentially causing samples to deviate from the learned data distribution \cite{habibian2024clockwork, ma2024deepcache, sabour2024align}. To mitigate this, it is crucial to periodically disable caching and refresh cached token updates, allowing the diffusion model to correct accumulated errors. \delete{Timesteps where caching is disabled are called Independent Steps (I-steps), while those where caching occurs are termed Prediction Steps (P-steps). } \revision{We refer to timesteps where caching is disabled as Independent Steps (I-steps) and those where caching occurs as Prediction Steps (P-steps).}

The core of our timestep-level token caching schedule is determining timesteps to be I-steps or P-steps. Inspired by recent advances in determining the optimal discretization of timesteps for diffusion models \cite{sabour2024align}, we propose using accumulated $\bm{\alpha}$ values to identify optimal I-steps. 
First, we collect the $\bm{\alpha}$ for a batch of $B=256$ samples and then average these across sample, block, and token dimensions to derive a timestep importance measure $\bar{\alpha}_t$. This measure is processed in exponential space and normalized to compute the importance distribution $\{v_t\}_{t=1}^T$ as follows:
\begin{align}
    v_t = \frac{\exp(\gamma_t)}{\sum_{i=t}^{T} \exp(\gamma_i)}, \quad
    \exp(\gamma_t) = \sum_{i=t}^{T} \exp(\beta \bar{\alpha}_i), 
    \label{eq:timestep_importance}
\end{align}
where $\beta$ is a smoothing hyperparameter. The percentiles of the distribution $\{v_t\}_t$ are selected uniformly as the final I-steps. These I-steps divide the timesteps into intervals, ensuring each interval contains approximately the same amount of smoothed accumulated cache decisions. This correlates with both local errors and token update similarities (\cref{eq:loss,eq:optimal_alpha}). These I-steps remain constant during subsequent evaluations.

One may be tempted to use local errors directly to determine I-steps (\eg, substituting $\alpha$ with estimated local errors in \cref{eq:timestep_importance}) because local errors reflect the accuracy of one-step predictions. However, local errors are often noisy with exaggeratedly large values, leading to suboptimal results. In contrast, the decision variables $\bm{\alpha}$ have two key advantages: 1) their values are within $[0, 1]$ and are smoother than unbounded local errors; 2) $\bm{\alpha}$ reflects the similarity of the token updates across timesteps (\cref{eq:optimal_alpha}), aligning well with the goal of TokenCache, \ie, reusing similar updates. Thus, using $\bm{\alpha}$ gives better performance.

\section{Experiments}

\subsection{Settings}

To evaluate the performance of TokenCache, we conducted experiments with widely recognized DiT variants: DiT-XL/2 \cite{peebles2023dit} with DDIM sampler \cite{song2020ddim} for class-conditioned image synthesis, PixArt-$\alpha$ \cite{chen2023pixart} with DPM-solver++ \cite{lu2022dpm} for text-to-image synthesis, and Open-Sora-Plan \cite{lin2024open} with Euler Ancestral Discrete Scheduler \cite{karras2022elucidating} for text-to-video synthesis. 

\begin{table}[b]
\vspace{-1.6em}
\caption{Inference hyperparameters. \#Full is the number of non-caching timesteps.}
\vspace{-.6em}
\centering
	\setlength\tabcolsep{6pt}
	\setlength{\extrarowheight}{2pt}
\begin{tabular}{l|c|c|c|c|c|c}
\hline
Model & \multicolumn{4}{c|}{DiT 256x256} & \multicolumn{2}{c}{DiT 512x512} \\ \hline
DDIM & 250 & 50 & 20 & 10 & 50 & 20 \\ \hline
\#Full & 50 & 12 & 9 & 6 & 21 & 10 \\ \hline
$\beta(\times10^3)$ & -6 & -8 & -4 & -2 & -4 & -2 \\ \hline
\end{tabular}
\label{tab:inference_hyperparameters}
\end{table}

\begin{table*}[ht]
\centering
\caption{\revision{\textbf{Class-conditioned image synthesis} on DiT-XL/2 with $256\times256$ and $512\times512$ resolutions, compared under similar speedup settings.} \textit{NFE} denotes the number of function evaluations, and \textit{Prec.} denotes precision. Latency is measured in seconds. \textbf{Bold} entries indicate the best results, and {\ul underlined} entries indicate the second best.}
\renewcommand\arraystretch{1.5}
\begin{tabular}{c|c|c|rrrrr|rrr}
\hline
\textbf{Method} & \textbf{Resolution} & \textbf{NFE} & \textbf{FID$\downarrow$} & \textbf{sFID$\downarrow$} & \textbf{IS$\uparrow$} & \textbf{Prec.$\uparrow$} & \textbf{Recall$\uparrow$} & \textbf{Latency$\downarrow$} & \textbf{TFLOPs$\downarrow$} & \textbf{Speedup$\uparrow$} \\ \hline
\rowcolor[HTML]{EFEFEF} 
\textbf{DiT-XL/2} & \textbf{256} & \textbf{250} & 2.09 & 4.58 & 243.24 & 0.81 & 0.61 & 21.57 & 59.34 & 1.00$\times$ \\ 
\textbf{DiT-XL/2} & 256 & 125 & {\ul 2.13} & {\ul 4.44} & {\ul 244.99} & {\ul 0.80} & {\ul 0.60} & 10.77 & 29.67 & 2.00$\times$ \\
\textbf{DiT-XL/2} & 256 & 84 & 2.18 & \textbf{4.36} & 243.90 & \textbf{0.81} & {\ul 0.60} & 7.25 & 19.94 & 2.98$\times$ \\
\textbf{FORA} & 256 & 250 & 2.82 & 6.04 & \textbf{253.96} & {\ul 0.80} & 0.58 & 8.60 & 20.02 & 2.96$\times$ \\
\textbf{ToCa} & 256 & 250 & 2.58 & 5.74 & 244.00 & \textbf{0.81} & 0.59 & 11.13 & 19.11 & 3.10$\times$ \\
\rowcolor[HTML]{ECF4FF} 
\textbf{Ours} & 256 & 250 & \textbf{2.07} & {\ul 4.44} & 244.29 & {\ul 0.80} & \textbf{0.62} & 10.51 & 19.70 & 3.01$\times$ \\ \hline
\rowcolor[HTML]{EFEFEF} 
\textbf{DiT-XL/2} & \textbf{256} & \textbf{50} & 2.25 & 4.32 & 240.24 & 0.80 & 0.59 & 4.32 & 11.86 & 1.00$\times$ \\ 
\textbf{DiT-XL/2} & 256 & 26 & 2.81 & \textbf{4.47} & 231.64 & {\ul 0.79} & \textbf{0.59} & 2.24 & 6.17 & 1.92$\times$ \\
\textbf{DiT-XL/2} & 256 & 21 & 3.34 & 4.79 & 224.73 & 0.78 & {\ul 0.58} & 1.79 & 4.98 & 2.38$\times$ \\
\textbf{FORA} & 256 & 50 & 40.82 & 30.25 & 71.17 & 0.44 & 0.54 & 2.04 & 4.99 & 2.37$\times$ \\
\textbf{Learning-to-Cache} & 256 & 50 & {\ul 2.61} & 4.73 & {\ul 238.63} & \textbf{0.80} & {\ul 0.58} & 2.36 & 5.94 & 2.00$\times$ \\
\textbf{ToCa} & 256 & 50 & 3.31 & 4.90 & 218.00 & 0.78 & \textbf{0.59} & 2.38 & 5.27 & 2.25$\times$ \\
\rowcolor[HTML]{ECF4FF} 
\textbf{Ours} & 256 & 50 & \textbf{2.33} & {\ul 4.72} & \textbf{265.30} & {\ul 0.79} & \textbf{0.59} & 2.27 & 4.94 & 2.40$\times$ \\ \hline
\rowcolor[HTML]{EFEFEF} 
\textbf{DiT-XL/2} & 256 & \textbf{20} & 3.51 & 4.93 & 222.80 & 0.78 & 0.59 & 1.72 & 4.74 & 1.00$\times$ \\ 
\textbf{DiT-XL/2} & 256 & 14 & {\ul 5.68} & 6.55 & {\ul 201.23} & {\ul 0.75} & {\ul 0.56} & 1.20 & 3.32 & 1.50$\times$ \\
\textbf{DiT-XL/2} & 256 & 12 & 7.80 & 8.04 & 184.68 & 0.72 & 0.54 & 1.01 & 2.84 & 1.67$\times$ \\
\textbf{Learning-to-Cache} & 256 & 20 & 5.82 & {\ul 6.05} & 195.04 & {\ul 0.75} & 0.55 & 1.07 & 2.78 & 1.71$\times$ \\
\textbf{ToCa} & 256 & 20 & 8.02 & 7.75 & 179.33 & 0.72 & 0.54 & 1.17 & 2.76 & 1.82$\times$ \\
\rowcolor[HTML]{ECF4FF} 
\textbf{Ours} & 256 & 20 & \textbf{3.41} & \textbf{4.91} & \textbf{232.47} & \textbf{0.78} & \textbf{0.58} & 1.15 & 2.74 & 1.73$\times$ \\ \hline
\rowcolor[HTML]{EFEFEF} 
\textbf{DiT-XL/2} & \textbf{256} & \textbf{10} & 12.30 & 11.23 & 158.08 & 0.67 & 0.52 & 0.85 & 2.37 & 1.00$\times$ \\ 
\textbf{DiT-XL/2} & 256 & 8 & {\ul 22.94} & {\ul 19.19} & {\ul 120.29} & {\ul 0.57} & {\ul 0.48} & 0.68 & 1.89 & 1.25$\times$ \\
\textbf{DiT-XL/2} & 256 & 7 & 33.77 & 27.60 & 92.68 & 0.48 & 0.46 & 0.62 & 1.66 & 1.43$\times$ \\
\textbf{ToCa} & 256 & 10 & 57.20 & 32.12 & 58.46 & 0.36 & 0.50 & 0.67 & 1.71 & 1.39$\times$ \\
\rowcolor[HTML]{ECF4FF} 
\textbf{Ours} & 256 & 10 & \textbf{13.99} & \textbf{10.68} & \textbf{148.79} & \textbf{0.60} & \textbf{0.54} & 0.65 & 1.64 & 1.45$\times$ \\ \hline
\rowcolor[HTML]{EFEFEF} 
\textbf{DiT-XL/2} & \textbf{512} & \textbf{50} & 3.43 & 5.31 & 204.93 & 0.81 & 0.55 & 23.01 & 52.48 & 1.00$\times$ \\ 
\textbf{DiT-XL/2} & 512 & 28 & 4.02 & {\ul 5.78} & 196.14 & 0.80 & \textbf{0.54} & 12.96 & 29.38 & 1.79$\times$ \\
\textbf{Learning-to-Cache} & 512 & 50 & {\ul 3.75} & {\ul5.73} & {\ul 198.01} & \textbf{0.83} & \textbf{0.54} & 14.32 & 31.17 & 1.68$\times$ \\
\textbf{ToCa} & 512 & 50 & 4.72 & 5.90 &  194.44 & \textbf{0.83} & {\ul0.53} & 13.77 & 28.86 & 1.82$\times$ \\
\rowcolor[HTML]{ECF4FF} 
\textbf{Ours} & 512 & 50 & \textbf{3.44} & \textbf{5.57} & \textbf{207.31} & {\ul 0.82} & \textbf{0.54} & 12.92 & 28.89 & 1.82$\times$ \\ \hline
\rowcolor[HTML]{EFEFEF} 
\textbf{DiT-XL/2} & \textbf{512} & \textbf{20} & 5.17 & 5.74 & 183.97 & 0.82 & 0.56 & 9.20 & 20.99 & 1.00$\times$ \\ 
\textbf{DiT-XL/2} & 512 & 12 & 10.44 & 8.80 & 149.79 & 0.78 & {\ul0.52} & 5.56 & 12.59 & 1.67$\times$ \\
\textbf{Learning-to-Cache} & 512 & 20 & {\ul 7.38} & {\ul 7.17} & {\ul 168.99} & {\ul 0.81} & {\ul 0.52} & 6.41 & 14.25 & 1.47$\times$ \\
\textbf{ToCa} & 512 & 20 & {13.50} & {\ul 11.18} & {\ul 134.28} & {\ul 0.74} & \textbf{0.53} & 6.27 & 13.57 & 1.55$\times$ \\
\rowcolor[HTML]{ECF4FF} 
\textbf{Ours} & 512 & 20 & \textbf{5.43} & \textbf{6.21} & \textbf{185.64} & \textbf{0.83} &  0.50 & 5.76 & 12.86 & 1.63$\times$ \\ \hline
\end{tabular}

\label{tab:performance_dit_256}
\end{table*}

For DiT, we used the ImageNet dataset \cite{russakovsky2015imagenet}, a benchmark in image generation and classification, chosen for its diverse object classes and high-quality images. For PixArt-$\alpha$, we performed experiments on the MSCOCO-2017 dataset \cite{lin2014microsoft} and PartiPrompts \cite{yu2022scaling} for evaluation.
For Open-Sora-Plan, we leveraged the VBench framework~\cite{huang2023vbench}, 
a benchmark designed for video generation evaluation. 
\revision{VBench\,(higher is better) provides an aggregate measure of video quality, 
emphasizing temporal consistency, motion accuracy, and overall fidelity.} 
VBench was chosen for its diverse and comprehensive set of 950 benchmark prompts, 
covering various aspects of video generation quality.

\revision{We adopted several widely used quantitative metrics to evaluate model performance. 
\textbf{Fr\'echet Inception Distance (FID)}~\cite{heusel2017fid} 
\revision{(lower is better)} measures the distributional discrepancy between features of real and generated data, 
reflecting how closely the generated distribution aligns with the real one (i.e., overall fidelity). 
\textbf{Scaled FID (sFID)}~\cite{szegedy2016sfid} 
\revision{(lower is better)} is a scale-corrected variant of FID that normalizes feature covariances, 
reducing estimation bias and improving stability under limited samples. 
\textbf{Inception Score (IS)}~\cite{salimans2016inception} 
\revision{(higher is better)} evaluates both diversity and classifiability by measuring 
how confidently and diversely a pretrained classifier recognizes generated samples. 
\textbf{Precision and Recall}~\cite{kynkaanniemi2019precision} 
\revision{(higher precision = fidelity; higher recall = coverage/diversity)} 
quantify the trade-off between realism and diversity without the Gaussian assumption of FID, 
providing complementary insights into distributional fidelity and coverage. 
\textbf{CLIP Score}~\cite{hessel2021clipscore} 
\revision{(higher is better)} assesses text--image alignment via cosine similarity between text and image embeddings, 
capturing semantic consistency. 
\textbf{Image Reward}~\cite{xu2023imagereward} (higher is better) evaluates perceptual image quality and text–image alignment based on a reward model trained on large-scale human preference data. 
Unless otherwise noted, we used the same prompt set, sample size $N$, resolution, guidance scale, number of steps, 
and random seeds across all methods. 
We also equalized multiply–accumulate operations (MACs) per step to match overall FLOPs, 
ensuring fair comparisons of quality and efficiency.}


\begin{figure*}
    \centering
    \includegraphics[width=\linewidth]{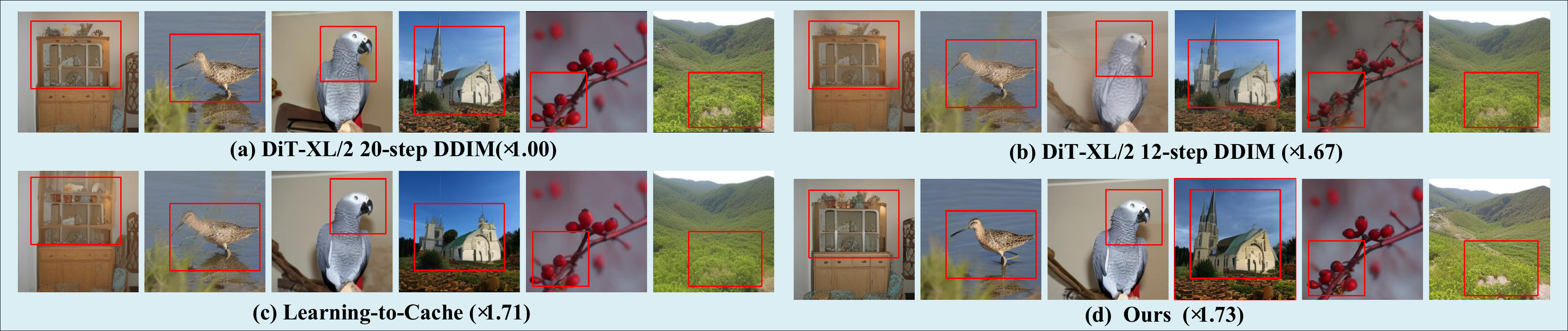}
    \caption{\revision{\textbf{Class-conditioned image synthesis} on DiT-XL/2 at $256\times256$ resolution.} Visual comparison among DiT-XL/2 (20 steps), DiT-XL/2 (12 steps), Learning-to-Cache, and our method.}
    \label{fig:vis_compare}
    \vspace{-.6em}
\end{figure*}%

\subsection{Implementation Details}

This section provides the details of the implementation. 
For image sampling, we employed the pretrained VAE\footnote{\url{https://huggingface.co/stabilityai/sd-vae-ft-ema}} from Stable Diffusion \cite{rombach2022high}.  For video sampling, we followed the settings in Open-Sora-Plan \cite{lin2024open}.
We used the AdamW optimizer \cite{loshchilov2017decoupled} for finetuning, where $\beta_1=0.9$ and $\beta_2=0.999$. The learning rate was \(1 \times 10^{-5}\) for 40K steps. After that, we froze the parameters of the Cache Predictor and continued to fine-tune the LoRA parameters of the last \(K\) layers with a learning rate of \(1 \times 10^{-4}\).
The default hyperparameters were $\lambda=0.0001$ for loss computation in \cref{eq:loss} and $K=4$ for LoRA finetuning. The cache ratio was $r=0.9$.  \delete{Generation speed was measured by latency and FLOPs with a batch size of $8$.} \revision{We measured the generation speed by latency and FLOPs using a batch size of $8$.}
\cref{tab:inference_hyperparameters} listed the inference hyperparameters for the DiT-XL/2 model. PixArt-$\alpha$ used the same hyperparameters as DiT at $256\times256$ resolution. Open-Sora-Plan used a full timestep of 18, $\beta=-8\times10^3$.
For DiT, we fine-tuned the models on the ImageNet dataset and generated 50,000 samples for benchmarking using a fixed random seed of $0$. 
For PixArt-$\alpha$, we fine-tuned the model on the MSCOCO-2017 dataset; one image was sampled for each prompt in the validation set of MSCOCO-2017 and PartiPrompts for benchmarking.
For Open-Sora-Plan, we fine-tuned the model on the Mixkit dataset; we generated 5 videos for each benchmark prompt under different random seeds, resulting in a total of 4,750 videos. The generated videos were comprehensively evaluated across 16 aspects proposed in VBench. Besides, all experiments were conducted on 8 NVIDIA A100 GPUs.

\subsection{Performance Comparisons}

This section compares our TokenCache with reduced step sampling and existing cache-based acceleration methods.
To establish a baseline for performance comparisons, we first describe the configuration for normal inference without caching. \delete{In this configuration, no token caching is applied, and each step of the diffusion process is computed in its entirety, representing the standard approach for diffusion-based image generation.}
\revision{In this configuration, we disable token caching and compute each step of the diffusion process in its entirety, representing the standard approach for diffusion-based image generation.}
We also provide a baseline with comparable inference speed for reference.

\begin{table*}[t]
\centering
\caption{\revision{\textbf{Text-to-image synthesis} on PixArt-$\alpha$ with $256\times256$ resolution, compared under similar speedup settings.} \textit{NFE} denotes the number of function evaluations, and \textit{CLIP} denotes the CLIP score. Latency is measured in seconds. \textbf{Bold} entries indicate the best results, and {\ul underlined} entries indicate the second best.}

\renewcommand\arraystretch{1.5}
\begin{tabular}{c|c|cc|c|ccc}
\hline
 &  & \multicolumn{2}{c|}{\textbf{MSCOCO-2017}} & \textbf{PartiPrompts} &  &  &  \\
\multirow{-2}{*}{\textbf{Method}} & \multirow{-2}{*}{\textbf{NFE}} & \textbf{FID$\downarrow$} & \textbf{CLIP$\uparrow$} & \textbf{CLIP$\uparrow$} & \multirow{-2}{*}{\textbf{Latency$\downarrow$}} & \multirow{-2}{*}{\textbf{TFLOPs$\downarrow$}} & \multirow{-2}{*}{\textbf{Speedup$\uparrow$}} \\ \hline
\rowcolor[HTML]{EFEFEF} 
\textbf{PixArt-$\alpha$} & \textbf{20} & 40.00 & 31.11 & 31.69 & 2.394 & 5.74 & 1.00$\times$ \\ 
\textbf{PixArt-$\alpha$} & 12 & 43.67 & 30.79 & 31.22 & 1.511 & 3.44 & 1.67$\times$ \\
\textbf{FORA} & 20 & {\ul 39.92} & {\ul 31.10} & {\ul 31.48} & 1.590 & 3.15 & 1.82$\times$ \\
\textbf{ToCa} & 20 & 40.81 & 31.01 & 31.45 & 2.202 & 2.96 & 1.94$\times$ \\
\rowcolor[HTML]{ECF4FF} 
\textbf{Ours} & 20 & \textbf{39.60} & \textbf{31.15} & \textbf{31.62} & 1.502 & 2.95 & 1.95$\times$ \\ \hline
\end{tabular}

\label{tab:performance_pix_256}
\end{table*}

\begin{table*}[h]
\centering
\caption{\revision{\textbf{Text-to-image synthesis} on PixArt-$\alpha$ with $1024\times1024$ resolution, compared under similar speedup settings.} \textit{NFE} denotes the number of function evaluations, and \textit{CLIP} denotes the CLIP score. Latency is measured in seconds. \textbf{Bold} entries indicate the best results, and {\ul underlined} entries indicate the second best.}
\renewcommand\arraystretch{1.5}
\begin{tabular}{c|c|cc|ccc}
\hline
\textbf{Method} & \textbf{NFE} & \textbf{Image Reward$\uparrow$} & \textbf{CLIP$\uparrow$} & \textbf{Latency$\downarrow$} & \textbf{TFLOPs$\downarrow$} & \textbf{Speedup$\uparrow$} \\ \hline
\rowcolor[HTML]{EFEFEF} 
\textbf{PixArt-$\alpha$} & \textbf{50} & 1.0127 & 32.13 & 42.15 & 324.86 & 1.00$\times$ \\ 
\textbf{PixArt-$\alpha$} & 25 & 1.0113 & 32.00 & 21.07 &  162.43 &  2.00$\times$ \\
\textbf{FORA} & 50 &  1.0109  & 31.98  & 22.46  & 168.93  & 1.92 $\times$ \\
\textbf{ToCa} & 50 & {\ul 1.0117} & {\ul 32.04} & 44.19 & 165.37  & 1.96$\times$ \\
\rowcolor[HTML]{ECF4FF} 
\textbf{Ours} & 50 & \textbf{1.0121} & \textbf{32.09} & 23.50 & 164.86 & 1.97$\times$ \\ \hline
\end{tabular}

\label{tab:performance_pix_1024}
\end{table*}

\begin{figure*}[pht]
    \centering
    \includegraphics[width=0.75\textwidth]{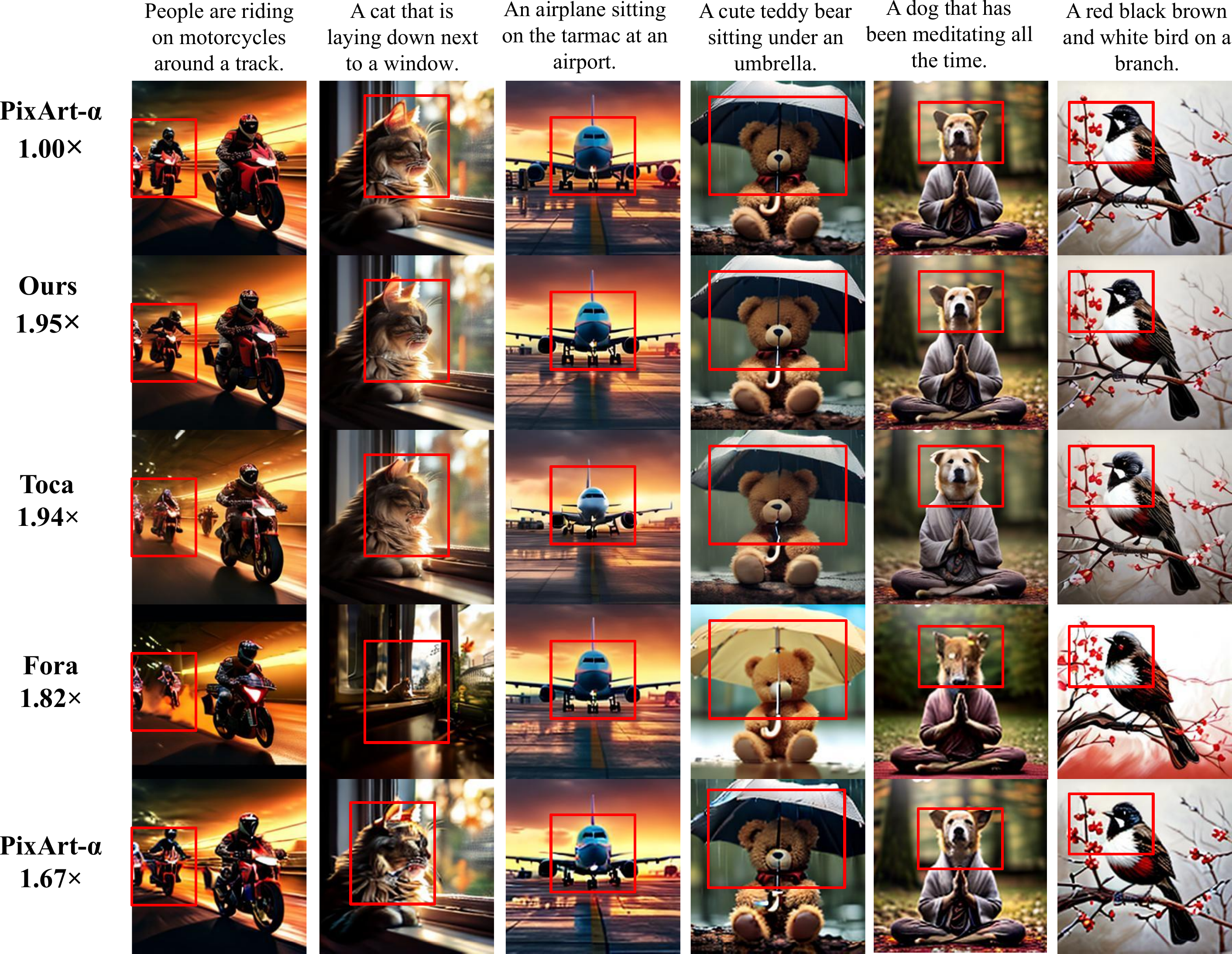}
    \caption{\revision{\textbf{Text-to-image synthesis} on PixArt-$\alpha$ at $256\times256$ resolution.} Visual comparison among PixArt-$\alpha$ (20 steps), PixArt-$\alpha$ (12 steps), ToCa, FORA, and our method.}
    \label{fig:pixart_vis}
\end{figure*}

\begin{table*}[t]
\centering
\caption{\revision{\textbf{Text-to-video synthesis} on Open-Sora-Plan with $256\times256$ resolution, compared under similar speedup settings.} \textit{NFE} denotes the number of function evaluations, and \textit{VBench} denotes the VBench score. Latency is measured in seconds. \textbf{Bold} entries indicate the best results, and {\ul underlined} entries indicate the second best.}

\renewcommand\arraystretch{1.5}
\begin{tabular}{c|c|c|ccc}
\hline
\textbf{Method} & \textbf{NFE} & \textbf{VBench$\uparrow$} & \textbf{Latency$\downarrow$} & \textbf{TFLOPs$\downarrow$} & \textbf{Speedup$\uparrow$} \\ \hline
\rowcolor[HTML]{EFEFEF} 
\textbf{Open-Sora-Plan} & 50 &  73.15 & 29.23  & 1717.49 & 1.00$\times$ \\ 
\textbf{Open-Sora-Plan} & 25 & 71.38 & 14.73 & 858.74 & 2.00$\times$ \\  
\textbf{FORA} & 50 &  70.96 & 15.54 & 790.10 & 2.17$\times$ \\  
\textbf{ToCa} & 50 & 71.92 & 14.54 & 764.60 & 2.25$\times$ \\  
\rowcolor[HTML]{ECF4FF}
\textbf{Ours} & 50 &  72.36 & 14.48 & 743.51 & 2.31$\times$ \\ \hline
\end{tabular}

\label{tab:performance_Open_sora_plan_256}
\end{table*}

\begin{figure*}[pht]
    \centering
    \includegraphics[width=0.75\textwidth]{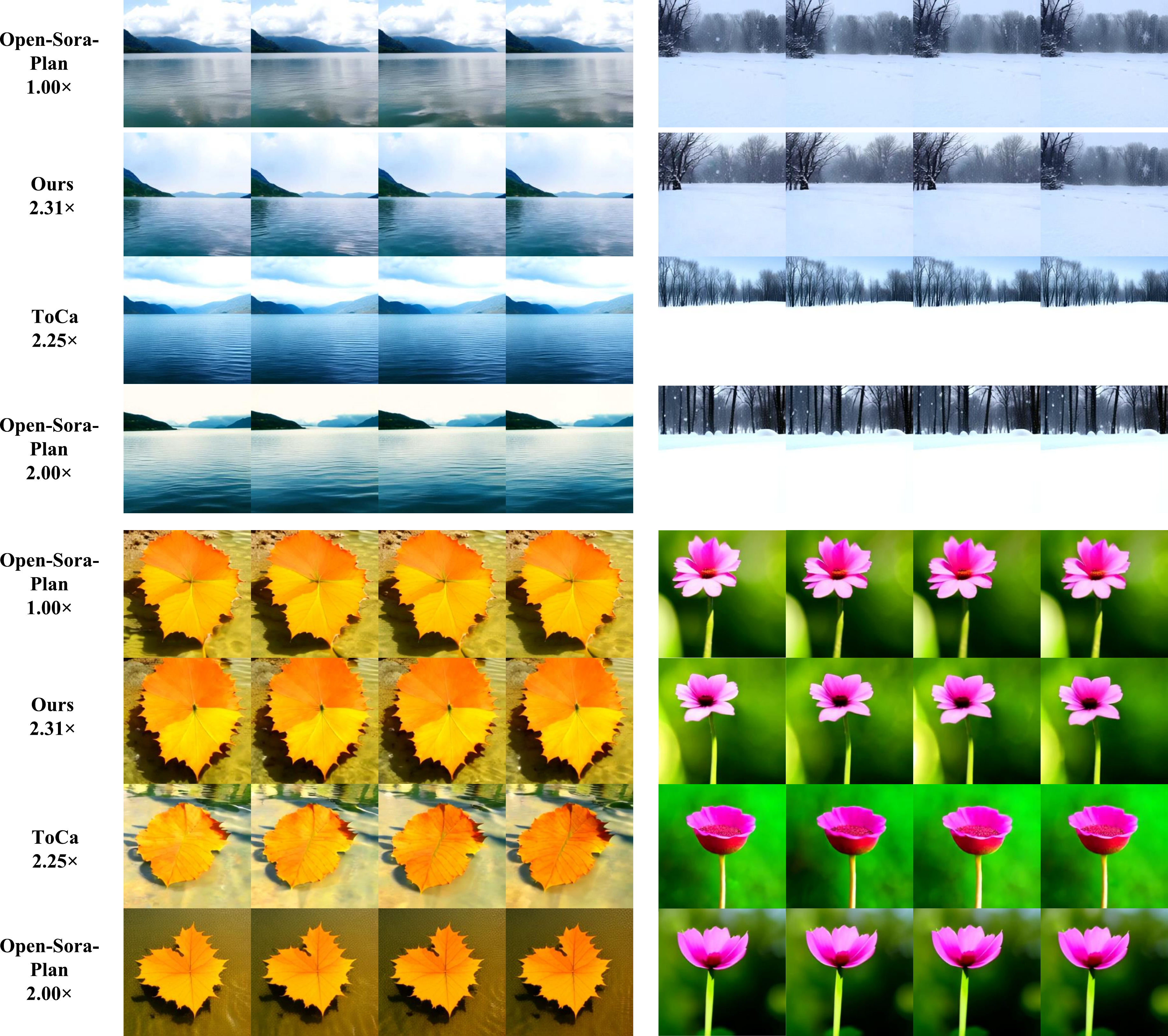}
    \caption{\revision{\textbf{Text-to-video synthesis} on Open-Sora-Plan at $256\times256$ resolution.} Visual comparison among Open-Sora-Plan (50 steps), Open-Sora-Plan (25 steps), ToCa, and our method.}
    \label{fig:open-sora-plan_vis}
\end{figure*}

\vspace{0.5em}
\noindent\textbf{Class-conditional synthesis.}
\cref{tab:performance_dit_256} presented a comparison of our method on DiT at a $256\times256$ resolution with other acceleration techniques: FORA \cite{selvaraju2024fora}, Learning-to-Cache \cite{ma2024learning}, and ToCa \cite{toca}. When matched to the same acceleration ratio, our method achieved the best FID and a higher IS among the compared methods, demonstrating its ability to balance speed and generation quality across both high and low sampling steps. As shown in \cref{fig:vis_compare}, our approach maintained consistent generation results even under a high acceleration ratio with fewer sampling steps, highlighting its robustness and effectiveness.
\revision{To make the visual advantages clearer, we annotated the improved regions in Fig. 5 with red bounding boxes. They clearly marked areas such as bookshelf structures, building outlines, and hillside protrusions where our method better preserved semantic consistency.}

For the 250-step DDIM, we set $K=2$ to ensure high speedup, while $K=4$ was used in all other reports. In this setting, our method achieved a $3\times$ acceleration without a reduction in performance compared to the original model, surpassing both FORA and ToCa at the same acceleration. Furthermore, at lower sampling steps, our method continued to outperform both the baseline model and other acceleration techniques at equivalent acceleration ratios, demonstrating its effectiveness in maintaining superior performance even under more challenging conditions with fewer sampling steps.
As illustrated in \cref{fig:fig1}, FORA's performance significantly deteriorated with fewer sampling steps, so we excluded further comparisons at these settings.

\cref{tab:performance_dit_256} further compared our results at a $512\times512$ resolution, showing that our method was also effective for achieving higher speedups at this increased resolution. In addition, it could be seen from the results that methods like ToCa, which require a lot of parameter adjustments, could not adapt well to different timesteps, especially in the case of few-step sampling, showing a large performance degradation.

\vspace{0.5em}
\noindent\textbf{Text-to-image synthesis.}
\cref{tab:performance_pix_256} compared our method with other acceleration methods for text-to-image synthesis at $256\times256$ resolution. Note that PartiPrompts had no ground-truth image, so we only evaluated the CLIP score. Our method consistently beat DDIM inference with fewer sampling steps and the two previous cache-based methods, while our method achieved an impressive near $2\times$ speedup.
This demonstrated the effectiveness of our method in more practical settings. Note that the performance of PixArt-$\alpha$ significantly degraded with a decreased number of timesteps because the discretization error of sampling the diffusion process quickly rose when the number of sampling steps is very low. In such cases, acceleration methods that do not reduce sampling steps gained advantages. As shown in \cref{fig:pixart_vis}, our method could generate images that are closer to the text description. In comparison, FORA and ToCa yielded relatively lower-quality results under similar acceleration ratios. For instance, ToCa produced a warped umbrella in the "teddy bear" image, while FORA failed to maintain visual consistency across different scenarios.
\revision{
In particular, in the "cat" case, FORA reuses cached features across multiple timesteps without adapting to the evolving fine-grained details of the diffusion states.
This static reuse results in visible appearance deviations such as smoother textures or altered lighting compared with the full-step baseline, although the overall semantics of the prompt remained correct.
In contrast, token-level adaptive caching methods better preserved structural coherence and fine details under the same acceleration ratio.
}
\revision{Besides, our method better preserved fine details and structural fidelity of the original model, even under high acceleration, as illustrated in the regions marked by the red bounding boxes such as the second motorcycle, the cat’s fur, the bear’s umbrella, and the dog’s head.}

\revision{
We additionally evaluated our method on PixArt-$\alpha$ at $1024\times1024$ resolution.
Following the same evaluation protocol as ToCa, we conducted text-to-image generation on the \textit{PartiPrompts}~\cite{yu2022scaling} dataset containing 1,632 textual prompts.
As summarized in \cref{tab:performance_pix_1024}, we compared our method with the baseline PixArt-$\alpha$, FORA, and ToCa under similar computational budgets. 
Given the limitations of FID at high resolutions, particularly its reliance on Inception features computed from 299×299 inputs, we adopted
Image Reward \cite{xu2023imagereward} and CLIP score \cite{hessel2021clipscore} as primary metrics.
Image Reward reflects human-preference-aligned perceptual quality, while CLIP score measures text–image semantic consistency.
Under comparable speedup ratios, our approach achieved the highest Image Reward and CLIP scores, 
demonstrating that the proposed caching mechanism maintains superior generation quality and semantic alignment even at high resolutions.
}

\vspace{0.5em}
\noindent\textbf{Text-to-Video synthesis.}
\cref{tab:performance_Open_sora_plan_256} compared methods on the Open-Sora-Plan dataset for text-to-video synthesis at $256\times256$ resolution. With similar acceleration ratios, our method obtained higher VBench scores compared to ToCa and FORA, reflecting its effectiveness in balancing quality and efficiency. Moreover, compared to the baseline Open-Sora-Plan, our approach achieved significant reductions in computational costs and latency, resulting in the highest overall speedup (2.18 $\times$), which demonstrated the scalability of our method on video generation tasks.
As shown in \cref{fig:open-sora-plan_vis}, our method could better maintain generation quality and consistency compared to ToCa and the original Open-Sora-Plan model with fewer steps. For example, when describing a scene of fallen leaves in water, our method could generate shadows and leaf shapes in a more detailed manner.

\begin{figure*}[t]
    \centering
    
    \includegraphics[width=\linewidth]{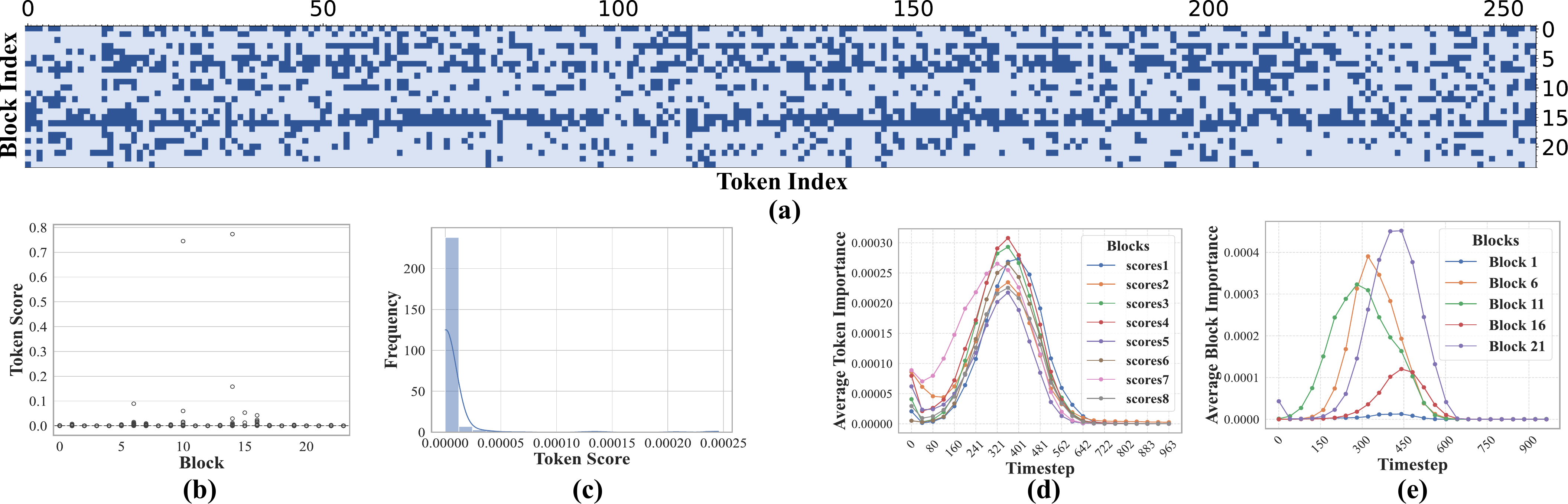}
    \caption{Analysis of Cache Predictor. (a) shows the cache status of each block at the cache ratio of 0.7. The \textcolor[rgb]{0.76, 0.80, 0.90}{light blue} is cached and the \textcolor{blue}{dark blue} is retained. (b) and (c) show the distribution of cache decision variables under different blocks. (d) and (e) show the importance of different tokens and blocks across timesteps, respectively.}
    \label{fig:score}
\vspace{-.6em}
\end{figure*}

\begin{figure}[t]
    \centering
    \includegraphics[width=3.5in]{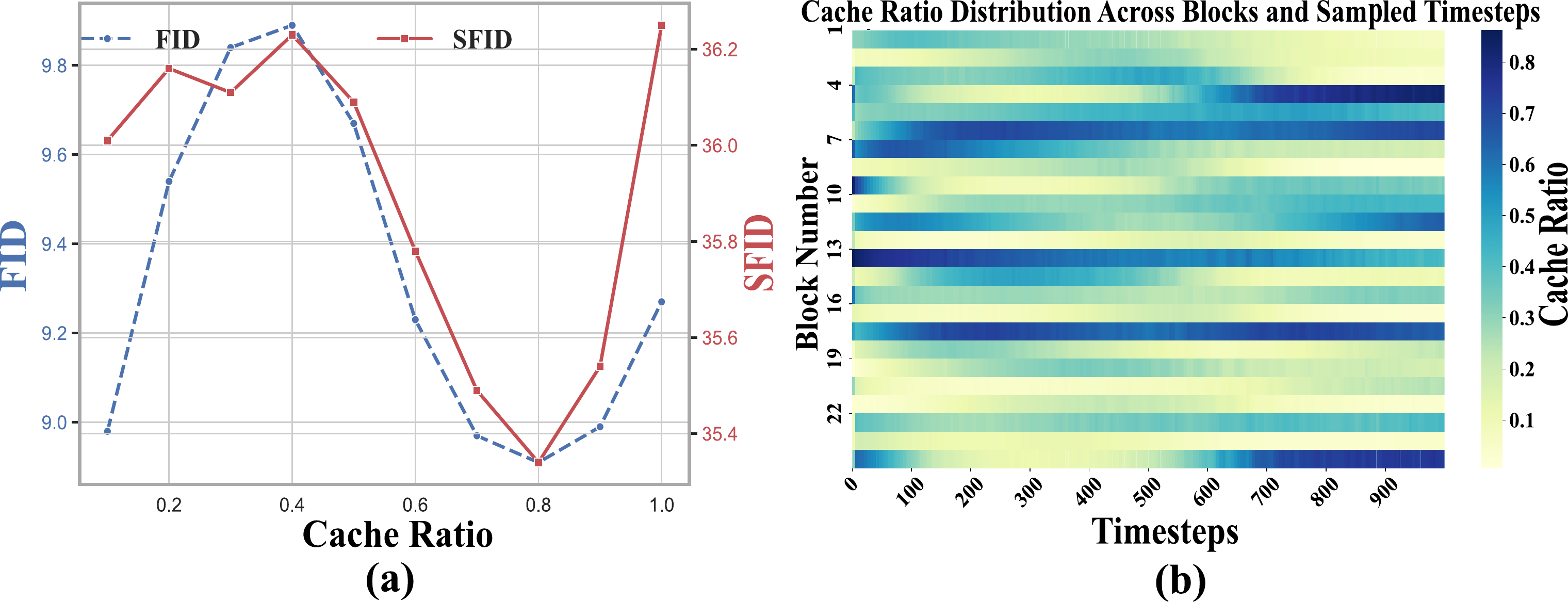}
    \caption{Analysis of the global cache ratio on 50-step DDIM. (a) Effect of varying the global cache ratio on performance metrics (FID and sFID). (b) Adaptive selection of cache ratios for each block at different timesteps when the global cache ratio is set to 0.7. The cache ratio shown in the figure corresponds to block-level caching.}
    \label{fig:cache}
\vspace{-.6em}
\end{figure}

\subsection{Ablation Study}

This section ablates the design choices of the TokenCache. 

\subsubsection{Performance Impact of Cache Prediction}

This section compares our Cache Predictor with different cache-predicting strategies, including random caching and ToCa's caching \cite{toca}. We also included the result of a token-pruning method, ToMe \cite{bolya2022token}. Specifically, we adapted ToMe to our framework by reusing the tokens that were decided to be merged by ToMe. Note that directly applying ToMe without caching led to poor performance.
All compared methods had similar computational costs.
The results are summarized in \cref{tab:ap_cache}. Our Cache Predictor achieved the best generation quality measured in FID and CLIP Score in all datasets. This demonstrated the effectiveness of our Cache Predictor. An additional benefit of our approach is that it requires less hyperparameter tuning than the compared methods.
\cref{fig:score} shows the distribution of the predictions of our Cache Predictor. As could be seen from the figure, our Cache Predictor gave sparse and consistent predictions for images, highlighting its effectiveness. 
As illustrated in \cref{fig:score}(a), certain blocks tend to cache more tokens, exhibiting a non-uniform distribution pattern that would be challenging to capture with handcrafted strategies. 
\cref{fig:score}(b) and \cref{fig:score}(c) further revealed that the cache decision variables are sparsely distributed across different blocks. This sparsity allowed our method to achieve higher caching ratios without significantly compromising generation quality.
Furthermore, \cref{fig:score}(d) shows that the predicted cache decision variables exhibited similar trends across different batch sizes at the same timestep, demonstrated the robustness of our Cache Predictor. In addition, \cref{fig:score}(e) revealed that each block followed a distinct evolution trend across timesteps. This observation supported the effectiveness of our block-level cache scheduling, which enabled each block to adjust its caching behavior based on its contextual importance. 
These observations above validated the effectiveness of our Cache Predictor and its ability to enhanced caching strategies through dynamic and context-aware decision-making. 

\revision{Although TokenCache is primarily designed for efficiency, it inherently maintains stable generation quality without semantic collapse. 
Even under aggressive caching ratios or long prediction horizons, the model self-corrected occasional mis-cached tokens through subsequent steps and periodic I-step refreshes.}

\begin{table}[htbp]
\centering
\setlength\tabcolsep{6pt}
\setlength{\extrarowheight}{2pt}
\caption{\revision{\textbf{Performance impact of cache prediction.}} Comparison of different cache-predicting strategies adapted to our framework.}

\begin{tabular}{c|cc|c}
\hline
\multirow{2}{*}{\textbf{Method}} & \multicolumn{2}{c|}{\textbf{MSCOCO-2017}} & \textbf{PartiPrompts} \\ \cline{2-4} 
 & \textbf{FID$\downarrow$} & \textbf{CLIP$\uparrow$} & \textbf{CLIP$\uparrow$} \\ \hline
Random & 40.26 & 30.60 & 31.26 \\
ToCa & 40.03 & 30.80 & 31.31 \\
ToMe & 40.05 & 30.88 & 31.37 \\
\textbf{Cache Predictor} & \textbf{39.60} & \textbf{31.15} & \textbf{31.62} \\ \hline
\end{tabular}
\label{tab:ap_cache}
\end{table}

\begin{table*}[h]
\centering
	\setlength\tabcolsep{6pt}
	\setlength{\extrarowheight}{1pt}
\caption{\revision{Effect of different global cache ratios on DiT-XL/2 (50-step DDIM, $256\times256$). 
Metrics include Inception Score (IS$\uparrow$), FID$\downarrow$, sFID$\downarrow$, 
Precision (Prec.$\uparrow$), Recall (Rec.$\uparrow$), TFLOPs$\downarrow$, and Speedup$\uparrow$.}}
\label{tab:cache_ratio_analysis_simplified}
\resizebox{.75\linewidth}{!}{
\revision{
\begin{tabular}{c|c c c c c c c}
\hline
Cache Ratio & IS$\uparrow$ & FID$\downarrow$ & sFID$\downarrow$ & Prec.$\uparrow$ & Rec.$\uparrow$ & TFLOPs$\downarrow$ & Speedup$\uparrow$ \\
\hline
0.9 & \textbf{267.68} & 8.99 & 35.54 & \textbf{0.811} & 0.7473 & \textbf{4.94} & \textbf{2.40} \\
0.8 & 265.48 & \textbf{8.91} & \textbf{35.34} & 0.7944 & 0.7540 & 5.88 & 2.02 \\
0.7 & 259.83 & 8.97 & 35.49 & 0.7854 & \textbf{0.7605} & 6.35 & 1.87 \\
0.6 & 244.08 & 9.23 & 35.78 & 0.7656 & 0.7709 & 7.30 & 1.62 \\
0.5 & 245.09 & 9.67 & 36.09 & 0.7394 & 0.7782 & 8.26 & 1.43 \\
\hline
\end{tabular}}
}
\end{table*}

\subsubsection{Analysis of the Cache Ratio}

We analyzed the impact of the cache ratio on 50-step DDIM.  \cref{fig:cache}(a) illustrated the effect of varying the global cache ratio on performance metrics (\eg, FID and sFID) for 50-step DDIM. Two contradicting factors existed when the global cache ratio increases: 1) Outdated Token Updates: More tokens from previous timesteps are reused, potentially introducing errors; 2) Dropping Uninformative Tokens: Less important tokens are cached, allowing the model to focus on critical updates. 
Initially, as the global cache ratio increases, cumulative errors per step rose, causing a performance drop. However, once the global cache ratio exceeds 0.4, the method retained only the most important tokens, leading to gradual improvements in FID and sFID. Beyond a ratio of 0.8, performance declined again due to the dropping of informative tokens. This phenomenon emphasizes the necessity of a fine-grained token caching strategy.

A high cache ratio could lead to significant information loss, degrading performance. Conversely, a low cache ratio is also problematic. The reason is that the trained cache predictor tended to assign low caching scores for most tokens, as they are mostly reusable. However, specifying a low cache ratio forced the cache predictor to preserve most tokens, which contradicts the prior it had learned and leads to a distribution shift.

\revision{
In addition to the analysis of FID and sFID in Fig.9, we evaluated the impact of the global cache ratio on a broader set of metrics, including Inception Score (IS), Precision, Recall, TFLOPs, and Speedup. Results were summarized in \cref{tab:cache_ratio_analysis_simplified}. 
As could be seen, a cache ratio of $0.8$ achieved the best fidelity (lowest FID and sFID in Fig.9) with balanced Precision and Recall. However, a ratio of $0.9$ achieved the highest Precision and the best speedup ($2.40\times$) with only a small drop in fidelity, offering a better trade-off for practical use. Ratios below $0.8$ reduced FLOPs but degraded IS and Precision noticeably. These results showed that 
$0.9$ provided a better balance between generation quality and inference efficiency, justifying its use as the default setting. 
}



\cref{fig:cache}(b) showed that when the global cache ratio is set to 0.7, our method adaptively allocates block-level cache ratios across different timesteps. Interestingly, some blocks consistently exhibited high cache ratios throughout the denoising process, indicating their limited contribution to final generation quality. This observation supported the effectiveness of block-level caching. 
Unlike static block-based methods, our approach introduces a fine-grained token-wise caching strategy within each block. This hierarchical design allowed for more flexible and precise update decisions, enabling our method to maintain superior performance even under higher acceleration ratios. By dynamically adjusting cache ratios at the global and block levels, while selectively caching tokens, TokenCache ensured a balance between speed and quality, enhancing overall inference efficiency. 

\revision{Furthermore, adjusting the cache ratio primarily affected Transformer performance by changing the balance between computational reuse and feature freshness, where higher ratios increased reuse but might slightly reduced feature update diversity, whereas lower ratios preserve more detailed updates at the cost of efficiency.
Benefiting from the adaptive Cache Predictor, periodic I/P-step refresh, and LoRA-refined terminal blocks, TokenCache effectively mitigated these effects, maintained stable representational capacity and high generation quality across all tested ratios.}


\begin{table}[htbp]
\vspace{-1em}
\caption{\revision{\textbf{Timestep schedules.}} Performance comparison of different timestep scheduling strategies.}

\setlength\tabcolsep{6pt}
\setlength{\extrarowheight}{2pt}
\centering
\begin{tabular}{l|c|c|c}
    \toprule
    \textbf{Timestep Strategy} & \textbf{IS$\uparrow$} & \textbf{FID$\downarrow$} & \textbf{sFID$\downarrow$} \\ \midrule
    \textbf{Reverse Selection} & 231.64 & 3.72 & 5.77 \\
    \textbf{Uniform Selection} & 250.20 & 2.51 & 5.36 \\ 
    \textbf{Ours} & 265.30 & 2.33 & 4.72 \\
    \bottomrule
\end{tabular}
\label{table:time_step_scheduling}
\end{table}

\subsubsection{Timestep Schedules}

We conducted an ablation study on timestep scheduling strategies. Our proposed timestep scheduling method is based on the timestep scores obtained from the Cache Predictor module. To validate the effectiveness of our approach, we performed a series of experiments. The first row in \cref{table:time_step_scheduling} showed results for a uniform timestep selection strategy, where cache time steps are applied evenly. The second row presented results for a reverse timestep selection strategy, in which the diffusion model uses larger time intervals during the later stages of generation. The results indicated that our adaptive timestep scheduling achieved a more efficient and balanced allocation of computational resources, enhancing both the quality and speed of the diffusion process.

\begin{table}[htbp]
\caption{\revision{\textbf{Performance impact of LoRA finetuning.} Results with different numbers of non-cached blocks $K$.}}

\centering
\begin{tabular}{l|c|c|c|c}
    \toprule
    \textbf{$K$} & \textbf{IS$\uparrow$} & \textbf{FID$\downarrow$} & \textbf{TFLOPs$\downarrow$}
    & \textbf{Speedup$\uparrow$}\\
    \midrule
    1 & 226.69 & 3.79 & 3.835 & 3.09$\times$ \\
    2 & 248.65 & 2.71 & 4.137 & 2.87$\times$ \\
    3 & 254.71 & 2.65 & 4.439 & 2.67$\times$ \\
    4 & 265.30 & 2.33 & 4.748  & 2.40$\times$ \\
    \bottomrule
\end{tabular}
\label{tab:LoRA}
\end{table}

\subsubsection{Performance Impact of LoRA Finetuning}

\begin{figure*}[pht]
    \centering
    \includegraphics[width=0.86\textwidth]{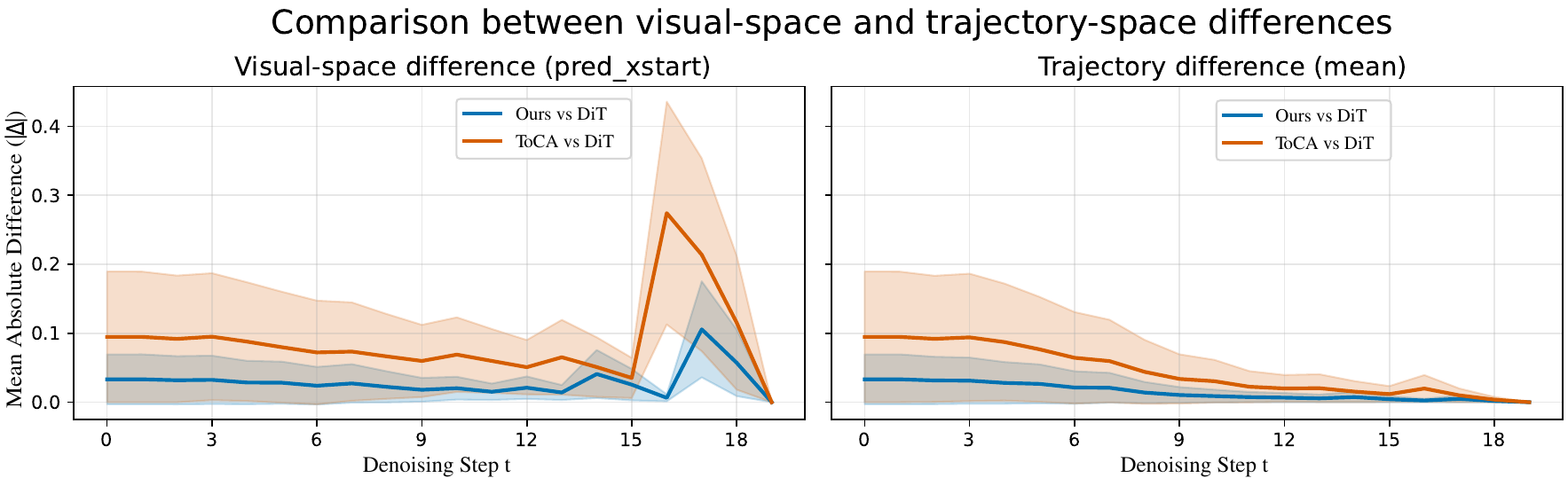}
    \caption{Per-step mean absolute difference  between DiT and different caching strategies. Left: visual-space deviation (pred\_xstart). Right: trajectory-space deviation (mean).}
    \label{fig:dit-dev}
    \vspace{-.6em}
\end{figure*}%

\begin{figure*}[ht]
    \centering
    \includegraphics[width=0.86\textwidth]{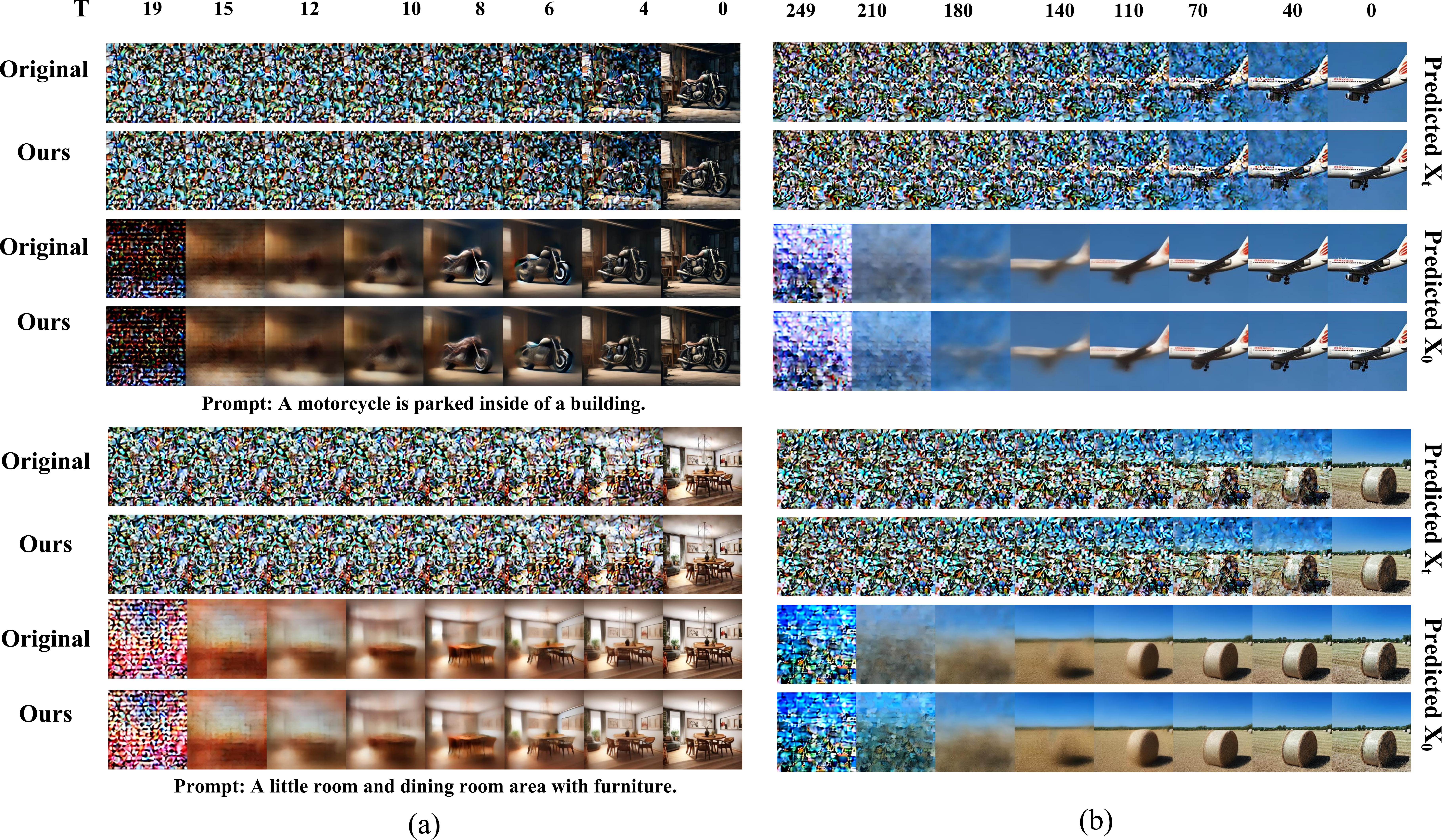}
    \caption{Comparative visualization of model outputs across different timesteps $T$. (a) shows PixArt-$\alpha$ results and (b) shows DiT results. The predicted $X_0$ refers to the denoised image estimated by the model at the current timestep. The predicted $X_t$ represents the posterior mean estimated by the model at the same timestep, which is used to sample the next latent in the reverse diffusion process.}
    \label{fig:vis1}
    \vspace{-.6em}
\end{figure*}



To ensure that the model maintains high generation quality even with a high cache ratio, we retained the final $K$ blocks as the "decoder" which did not perform caching and fine-tuned them using LoRA. \cref{tab:LoRA} presented the performance with different values of $K$ in terms of IS, FID, TFLOPs, and speedup. The results showed that increasing $K$ improved generation quality, as evidenced by higher IS and lower FID values. 
However, this improvement came at the cost of increased computational requirements, because the "decoder" does not perform caching. As the number of $K$ increases, the TFLOPs rose from 3.835 for a single layer to 4.748 for four layers, resulting in a gradual reduction in speedup from $3.09\times$ to $2.40\times$. We found that $K=4$ could well balance between generation quality and speedup.





\subsection{\revision{Deviation Analysis of Caching Strategies}}
\revision{
At each timestep $t$, we computed the per-pixel mean absolute difference (MAE) 
between the baseline DiT and each variant, and reported the mean $\pm$ standard deviation 
across a batch of images ($B=16$). 
Specifically, we evaluated two complementary deviation metrics:
(i) \textit{Visual-space difference}, which measures the per-pixel MAE between the predicted clean images $\hat{x}_0^{(t)}$ and the baseline outputs $x_0^{\text{baseline},(t)}$, 
reflecting the fidelity of the reconstructed images in the visual domain; and 
(ii) \textit{Trajectory difference}, which measures the MAE between the predicted posterior means $\hat{x}_t$ and the baseline latent states $x_t^{\text{baseline}}$, 
capturing the stability and consistency of the denoising trajectory over time.
As shown in Fig.~\ref{fig:dit-dev}, 
our proposed strategy (blue) consistently exhibited smaller deviations from the baseline DiT 
compared with the non-adaptive variant ToCa (orange), 
in both the predicted clean-image space (left) and the posterior-mean trajectory space (right). 
These results demonstrated that the joint design of adaptive cache selection and LoRA refinement 
effectively stabilized the denoising trajectory and better preserved reconstruction fidelity, 
highlighting the advantage of our strategy beyond simple cache adaptation.
}


\subsubsection{Visualizations of Token Caching Across Timesteps}
To further investigate the effectiveness of our token caching strategy, we conducted visualizations as part of the ablation experiments. Specifically, \cref{fig:vis1} provided a comparative visual analysis between the original model outputs and those produced by our token caching method across different timesteps. 
The comparison demonstrated that our method successfully retained critical information iteratively, maintaining the consistency and coherence of key features without significant deviation caused by the token pruning operations.


\section{Conclusion}
In this work, we introduce TokenCache, an innovative acceleration method designed for Diffusion Transformers (DiT). TokenCache leverage a Cache Predictor to selectively cache and reuse tokens, effectively addresses DiT's token-based computation structure to significantly enhance inference speed while preserving generation quality. By integrating token-level cache scheduling, block-level cache scheduling, and timestep scheduling strategies, TokenCache achieved an effective balance between computational efficiency and output fidelity. Extensive experiments on class-conditional image synthesis, text-to-image, and text-to-video generation demonstrate the effectiveness of the proposed TokenCache. It outperforms existing cache-based methods in both efficiency and quality, in both high and low sampling scenarios.
These results underscore the potential of fine-grained token caching as a powerful tool for advancing efficient diffusion-based generative modeling. 



\bibliographystyle{IEEEtran}
\bibliography{reference}

\vfill

\end{document}